\documentclass[journal]{IEEEtran}

\usepackage{pslatex, graphicx, amssymb, amsmath}

\usepackage{algorithmic}
\usepackage{algorithm}
\usepackage{subfigure}
\usepackage{amsthm}

\usepackage{tabularx}
\usepackage{pifont}
\usepackage{array}
\usepackage{ifthen}
\usepackage{xcolor}

\usepackage{csquotes}

\usepackage[hyphens]{url} 

\def\transpose{\perp}    
\def\learningrate{\alpha}
\def\weight{w}
\def\xstar{x_\star}
\def\ystar{y_\star}
\def\weightstar{w_\star}
\def\eigenvalueshessian{\mu}
\def\circone{\ding{172}}

\def\ball#1#2{B_{#1}(#2)}
\def\refb#1{(\ref{#1})}
\DeclareMathOperator{\diag}{diag}
\DeclareMathOperator{\Mat}{Mat}

\newif\ifextended
\extendedtrue

\newif\ifDeeinbinden
\Deeinbindenfalse

\newcolumntype{C}[1]{>{\centering\arraybackslash}p{#1}}

\newcommand{\bildgroesse}{.48\textwidth}

\newcommand{\norm}[1]{\left\lVert#1\right\rVert}
\newcommand{\N}{\mathbb{N}} 
\newcommand{\R}{\mathbb{R}} 
\newcommand{\C}{\mathbb{C}} 

\newtheorem{theorem}{Theorem}[section]
\newtheorem{lemma}[theorem]{Lemma}
\newtheorem{corollary}[theorem]{Corollary}

\begin{document}

\title{Local Convergence of Adaptive Gradient Descent Optimizers \thanks{This paper expands the results and presents proofs that are referenced in \cite{Bock.2019}.}}

\author{Sebastian~Bock~and~Martin~Georg~Wei{\ss}%
\thanks{S. Bock and M. Wei{\ss} are with the Department of Computer Science and Mathematics, OTH Regensburg, Pr\"ufeninger Str. 58, 93049 Regensburg, Germany}}

%



\maketitle

\begin{abstract}
Adaptive Moment Estimation (ADAM) is a very popular training algorithm for deep neural networks and belongs to the family of adaptive gradient descent optimizers.
However to the best of the authors knowledge no complete convergence analysis exists for ADAM. The contribution of this paper is a method for the local convergence analysis in batch mode for a deterministic fixed training set, which gives necessary conditions for the hyperparameters of the ADAM algorithm. Due to the local nature of the arguments the objective function can be non-convex but must be at least twice continuously differentiable. Then we apply this procedure to other adaptive gradient descent algorithms and show for most of them local convergence with hyperparameter bounds.
\end{abstract}


%
\IEEEpeerreviewmaketitle

\section{Introduction}
\IEEEPARstart{M}{any} problems in machine learning lead to a minimization problem in the weights 
of a neural network: 
Consider e.g. training data $(x_1, y_1), \ldots, (x_N, y_N)$ consisting of inputs $x_i$ and 
outputs $y_i$, and the task to determine a neural network
that has learned the relationship between inputs and outputs. This 
corresponds to a function $y = F( \weight , x )$, parametrized by the weights $\weight$,
which minimizes the average loss function 
$$
	f(\weight) = \frac 1 N \sum_{i=1}^N l(x_i,y_i,\weight) =:  \frac 1 N \sum_{i=1}^N f_i(\weight)
$$
over the training data. Typically the loss is built using
some norm for regression problems, e.g.
$
	l(x,y,\weight)=\frac 1 2 \norm {y - F(\weight, x)}_2^2
$, or using cross entropy for classification. Optimization algorithms construct a sequence
$\{\weight_t\}_{t\in\N_0}$ of weights starting from an initial value $\weight_0$, which under 
appropriate assumptions converges to some local minimum $\weightstar$ for general 
non-convex $f$. 
The most simple optimization algorithm for differentiable $f$ 
is gradient descent with the update
$\weight_{t+1} = \weight_t -\learningrate \nabla f(\weight_t)$ and a learning 
rate $\learningrate>0$. For convex $f$ conditions on the Lipschitz constant $L$ of $\nabla f$ 
guarantee convergence and give estimates for the rate of convergence, see \cite{Nesterov.2004}.
However $L$ is hard to get in practice, and choosing $\learningrate$ too big leads to
oscillatory behaviour. Besides it is well known from optimization that the gradient 
is not the only descent direction for $f$, neither is it optimal for finite step lengths,
see \cite{Nocedal.2006}, 
but computation of the Hessian is usually prohibitive. This has led to the development of a family of algorithms which compute moments of first order, 
that is approximate descent directions based on previous iterates of the gradient
like the initial momentum method \cite{Nesterov.2004}, 
as well as second order moments to control the componentwise scaling  and / or to adapt the 
learning rate in AdaGrad \cite{Duchi.2011} and ADAM \cite{Kingma.2015}. For the ADAM
variant studied in this paper see Algorithm 1.
More algorithms exist with variants like
batch mode vs. online or incremental mode 
-- using $\nabla f(\weight_t)$ in iteration $t$ vs. 
$ \nabla f_k(\weight_t)$ where $k$ iterates in a cyclic fashion over $1, \ldots, N$,
or deterministic vs. stochastic choice of the index $k$ for $\nabla f_k(\weight_t)$, 
stochastic assumptions for the observation of $\nabla f(\weight_t)$ or $\nabla f_k(\weight_t)$, 
and so on.

However for most of these algorithms only partial convergence results are known. 
The original proof of \cite{Kingma.2015} is wrong as has been noted by several authors,
see \cite{Bock.2017,Rubio.2017}.
Modifying the algorithm to AMSGrad, 
\cite{Reddi.2019} establishes bounds on $\norm{\nabla f(\weight_t)}$, 
similar to the results in \cite{Chen.2018} for a class of algorithms
called Incremental Generalized ADAM. 
Though none of the results shows convergence of the sequence $\{\weight_t\}_{t\in\N_0}$. 
Also the proofs are lengthy and hardly reuse results from each other, 
giving not much insight. General results from optimization cannot be used for several 
reasons: First, the moments usually cannot be proven to be a descent direction. Second, the 
learning rate cannot be shown to be a step size valid for the Wolfe conditions
for a line search, see \cite{Nocedal.2006}. The algorithm for the step taken in iteration
$t$ may explicitly contain the variable $t$ in much more complicated ways than 
$\frac{1}{t}$ in the Robbins-Monro approach \cite{Robbins.1951}.

The contribution of this paper is a generally applicable method, based on the 
theory of discrete time dynamical systems, which proves local convergence of ADAM. The results are purely qualitative because they hold
for learning rates sufficiently small, where "sufficiently small" is defined
in terms of the eigenvalues of the Hessian in the unknown minimum $\weightstar$. 

The outline of this paper is as follows: In Section \ref{sec:Fixed Point Analysis under Perturbation} we will discuss the preliminaries and the idea of the convergence prove, which will be discussed in Section \ref{sec:Convergence-Proof}. The generality of this method is shown in Section \ref{sec:Extension other optimiziers}, where we apply it to other optimizers like AdaDelta or AdaGrad. In numerical experiments in Section \ref{sec:Experiments} we show the heuristic evidence of our theoretical proof. Finally, in Section \ref{sec:Conclusion and Discussion} we summarize the paper and give an outlook on ways to expand the convergence proof.


\section{Fixed Point Analysis under Perturbation}
\label{sec:Fixed Point Analysis under Perturbation}

\subsection{Notation}
The symbol $\transpose$ denotes the transpose of a vector or matrix. We use the component-wise multiplication and division of vectors, as well as component-wise addition of vectors and scalar without any special notation. It should always be clear from the context which calculation method is used due to the size of the arguments.
For $f:\R^n\to\R$ the gradient and Hessian are written as $\nabla f$ and $\nabla^2 f$,
provided they exist. The vector spaces of functions which are once or twice continuously
 differentiable are denoted $C^1$ or $C^2$ respectively.
Throughout this paper we assume $f:\R^n\to\R^n$ at least continuously differentiable, 
twice continuously differentiable for some results. 
The open ball with radius $r$ around $x\in\R^n$ is denoted by $\ball r x = \{y\in\R^n: \norm{y-x}<r\}$ and $\norm x$ is any norm. We denote 
$\rho \left( A \right) = \max \{ |\lambda| | \lambda \text{ eigenvalue of } A \}$ the spectral radius of a matrix $A$ and $\diag \left(v \right) \in \R^{n\times n}$ describes a  matrix with components of $v \in \R^n$ on the diagonal.

\subsection{Related Work}
Stochastic gradient descent (SGD) becomes an effective method for optimizing noisy tasks. Especially in the area of neural networks SGD variants are partly responsible for big successes in the last years, see e.g. \cite{Krizhevsky.2012} or \cite{Hinton.2012b}. 
\\
Popular first-order SGD methods are AdaGrad \cite{Duchi.2011} and RMSProp \cite{Hinton.2012}. Kingma and Ba combine the advantages of these two methods and introduce the Adaptive Moment Estimation (ADAM) in  \cite{Kingma.2015} (see Algorithm \ref{alg:ADAM}). 
\begin{algorithm}
\caption{ADAM Optimizer}
\label{alg:ADAM}
\begin{algorithmic}[1]
\REQUIRE $\learningrate \in \R^+$, $\epsilon \in \R$ $\beta_1, \beta_2 \in (0,1)$, $\weight_0 \in \R^n$ and the function $f(\weight) \in C^2 \left( \R^n, \R \right)$
\STATE $m_0 = 0$, $v_0 = 0$, $t = 0$
\WHILE{$\weight$ not converged}
\STATE $m_{t+1} = \beta_1 m_t + (1-\beta_1) \nabla_w f(\weight_t) $ 
\STATE $v_{t+1} = \beta_2 v_t + (1- \beta_2) \nabla_w f(\weight_t)^2$
\STATE $\weight_{t+1} = \weight_t - \learningrate \frac{\sqrt{1-\beta_2^{t+1}}}{\left( 1-\beta_1^{t+1}\right)} \frac{m_{t+1}}{\sqrt{v_{t+1} + \epsilon^2}}$
\STATE $t = t+1$
\ENDWHILE
\end{algorithmic}
\end{algorithm}
Unfortunately, the ADAM optimizer is not always defined the same way. Kingma and Ba \cite{Kingma.2015} use $\sqrt{v} + \epsilon$ and bias correction.
The algorithms in \cite{Reddi.2019} and \cite{Chen.2018}  do not use an $\epsilon$ as well as \cite{Zou.2019}, but the latter initializes $v_0 = \epsilon$. All three apply bias correction in the learning rate $\learningrate_t$. We use  bias correction as described in \cite[Section 2]{Kingma.2015} and $\sqrt{v + \epsilon^2}$. $\sqrt{v + \epsilon^2}$ is more similar 
to the undisturbed $\sqrt{v}$ than $\sqrt{v}+\epsilon$ from \cite{Kingma.2015} or $\sqrt{v+\epsilon}$ from \cite{daSilva.2018}.

\begin{figure}[!h]
\center
\begin{subfigure}{}
\includegraphics[width=\bildgroesse]{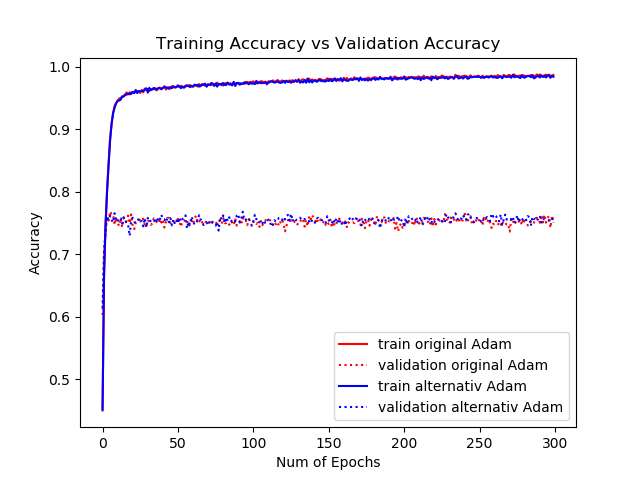}
\end{subfigure}
\begin{subfigure}{}
\includegraphics[width = \bildgroesse]{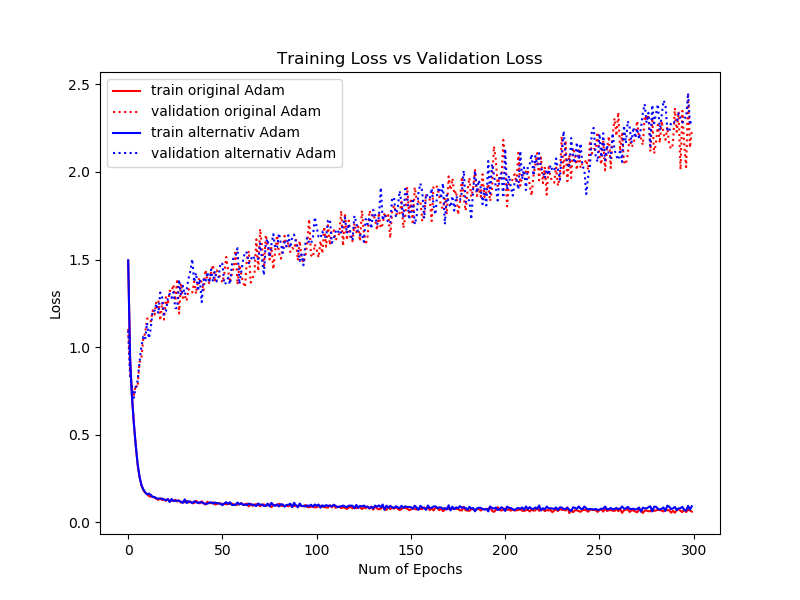}
\end{subfigure}
\caption{Train the fashion-mnist dataset (\cite{Xiao.2017}) with the two different ADAM methods}
\label{fig:Epsilon Experiment}
\end{figure}

The differences between the two possible usages of $\epsilon$ are minimal in their
effect on the evolution of loss and accuracy during learning 
(see Figure \ref{fig:Epsilon Experiment}\footnote{The experiment is programmed with Keras 2.2.4, Tensorflow 1.13.1 and Python 3.6.8} in the appendix) 

especially in the area around $v \approx 0$. 
The main aim by the introduction of $\epsilon$ -- avoiding division by 0 -- holds in both variants, but $\sqrt{v+\epsilon^2}$
gives the additional advantage of making the right hand side continuously differentiable
for $v \in[0,\infty)$ whereas $\sqrt v +\epsilon$ is not differentiable at $v=0$.
Differentiability will be essential in our proof. In Subsection \ref{subsec:ConnectionAdam} we will make the connection to ADAM as presented in \cite{Kingma.2015} and prove the local convergence, too. \\ \\
During the last years the ADAM Optimizer has become one of the most used optimization methods for training neural networks. Even if it is apparently working, there is, to the best of our knowledge, still no convergence proof for ADAM. The proof in the original paper \cite{Kingma.2015} was shown wrong, see \cite{Bock.2017}, \cite{Bock.2018} or \cite{Rubio.2017}. Reddi et. al. in \cite{Reddi.2019} present even a counter example and also introduce an improved method called AMSGrad. However in experiments AMSGrad does not show improvements at all (see \cite{Chen.2018} or \cite{Keskar.2017}). On the contrary, in some cases it ends up with worse accuracy than ADAM. 
Chen et al. \cite{Chen.2018} are showing for non-convex $f$ that $\min_{t = 1}^T  E [\norm{\nabla f(x_t)}^2] = O \left( \frac{s_1 \left( T \right)}{s_2 \left( T \right)}\right)$. 
With the assumption of $s_1 \left( T \right)$ growing slower than $s_2 \left( T \right)$, one reaches a minimum of $E[\norm{\nabla f \left( x_t \right)}^2]$ but without a guarantee of staying there. Barakat and Bianchi\cite{Barakat.20181} 
use a similar interpretation of the ADAM optimizer with a dynamical system viewpoint.  uses This approach uses a non-autonomous ordinary differential equation without
Lipschitz condition because of the term $\sqrt v + \epsilon$ instead of our non-autonomous system of difference equations. In our opinion, this choice makes the proof longer and more complicated.\\ \\
In the current work we present a convergence proof of the ADAM optimizer \cite{Kingma.2015} in a complete batch mode using all of the training data. With this assumption we can guarantee, that we are searching the same minimum $\weightstar$ in each time step $t$. Due to the local nature our proof does not assume 
the convexity of $f(\weight)$, thus we can guarantee local convergence even for non-convex settings. The hyperparameter setting is only restricted by 
\begin{align}
\label{eq:Our_Inequality}
\frac \learningrate {\epsilon} { \max_{i=1}^n \left(\eigenvalueshessian_i \right)}\left(1-\beta_1\right) < 2\beta_1 + 2
\end{align}
with $\eigenvalueshessian_i$ the $i$-th eigenvalue of the Hessian $\nabla^2 f(\weightstar)$. The counter example in \cite{Reddi.2019} or \cite{Luo.2019} does not affect our convergence proof, because we consider batch mode only. For example the incremental function in \cite{Reddi.2019} becomes a linear function in batch mode.\\ \\
All these considerations, we then apply to the most famous algorithms of the adaptive gradient decent family. We subsequently publish all resulting hyperparameter boundaries for each optimizer in Table \ref{tab:optimizer extension hyperparameter bounding}. It should be noted that we only focus on the supposedly most important algorithms in the area around ADAM. A table with all convergence statements for every algorithm in the adaptive gradient decent family would be desirable but not feasible due to the amount of algorithms.

\subsection{Idea}
We consider the learning algorithm from
the standpoint of dynamical systems and define a common state vector $x$ consisting
of the moments -- like $m$ and $v$ for ADAM -- and the weights, so we have 
$x =(m, v, w)$. Then the optimization can be written as an iteration 
$x_{t+1} = T(t, x_t)$ for some function $T: \N_0\times X\to X$ with $X\subset \R^{p}$, 
which defines a non-autonomous dynamical system. The function $f$ to be minimized
in the learning process, or rather its gradient, becomes a part of $T$.
If $f$ is at least continuously differentiable a local minimum gives the necessary 
condition $\nabla f(\weightstar)=0$. We show that this condition leads to a fixed point $\xstar$ of $T$, 
where the moments are all zero. We analyse the stability properties of this fixed
point and prove local asymptotic stability. This is done by considering a
time-variant iteration $T$ as the perturbation of a time-invariant iteration
$\bar T$ where Banach-like fixed point arguments can be applied. We use the second method of Lyapunov
for stability analysis where the vanishing moments simplify the computation and 
estimates for the eigenvalues.
Asymptotic stability however is equivalent to convergence of the iteration defined
$T$ to $\xstar$ for all $x_0$ sufficiently close to $\xstar$. The conditions needed
for the fixed point analysis and stability results require the learning rate to be 
sufficiently small.
Note that these results cannot be obtained directly from standard fixed point 
theorems for autonomous systems, because the iteration index $t$ enters the dynamics. 
Therefore also estimates of the eigenvalues depend on the iteration $t$, and even
a bound on the spectral radius uniform in $t$ does not give the convergence results
presented here: It is well known that $\rho(A)<1$ implies the existence of
a vector norm with induced matrix norm such that $\norm A<1$, but this norm depends on 
$A$. So $\rho(A_t)\leq c < 1$ for some $c$ for all $ t\in\N_0$ does not imply 
the existence of a \emph{single} norm such that $\norm{A_t}<1$ for all $t$.
We emphasize that the result is purely qualitative, giving no explicit guidance to the choice of the learning rates. 
The main advantage of our approach is the clearness of the proof, only computation of eigenvalues is needed once the iteration has been written in terms of $T$ and $\bar T$. These calculations are much more simple than the lengthy estimates in \cite{Kingma.2015}, \cite{Reddi.2019} and \cite{Chen.2018}.

We stress that local convergence result is all one can hope for: 
A global convergence proof cannot be obtained for most algorithms
including ADAM because a 2-cycle of the iteration exists for all values of 
hyperparameters for simple quadratic objective functions, see \cite{BockWeiss.2019c}.
\subsection{Preliminaries}
We recall some standard definitions and facts from the theory of difference equations
and discrete time dynamical systems, see e.g. \cite[Definition 5.4.1]{Agarwal.2000}
or \cite{Kelley.2001}. 
Consider $T:\N_0\times M\to M$ with $M\subset \R^p$ which defines a 
non-autonomous dynamical system by the iteration 
\begin{equation}
\label{eqDiscreteTimeVariant}
 x_{t+1} = T (t, x_t), \quad t\in \N_0, x_0 \in M
\end{equation}
with solutions 
$x:\N_0\to M$, $t\mapsto x_t$ depending on the initial value $x_0$. 
We use the notations $x_t = x(t; x_0)$ and $x=x(\cdot; x_0)$
to emphasize the dependence of solutions on the initial value if necessary. 
We always use the initial time $t_0=0$.

Autonomous systems constitute the special case where $T$ does not depend on $t$, 
so we can abbreviate to $\bar T: M\to M$ and write
\begin{equation}
\label{eq:DiscreteTimeInvariant}
 x_{t+1} = \bar T (x_t), \quad t\in \N_0, x_0\in M
\end{equation}
A point $\xstar\in M$ is called \emph{equilibrium} or \emph{fixed point}
if $T(t,\xstar) = \xstar$ for all
$t\in\N_0$, so the constant function $x_t = \xstar$ for all $t\in\N_0$ is a solution
of \refb{eqDiscreteTimeVariant}. In the following the asterisk will always denote
equilibria or their components.
 Consider a solution $x= x(\cdot; x_0)$ of \refb{eqDiscreteTimeVariant}.
$x$ is called \emph{stable}, 
if for each $\varepsilon>0$ there exists $\delta =\delta(\varepsilon)$ such that any 
solution $\tilde x = \tilde x(\cdot; \tilde x_0)$ of \refb{eqDiscreteTimeVariant} with 
$\norm{\tilde x_0-x_0}< \delta$ fulfills $\norm{\tilde x_t-x_t} < \varepsilon$
for all $t\in\N_0$. 

$x$ is called \emph{attractive} if there exists $\delta>0$ such that any solution
$\tilde x$ with $\norm{\tilde x_0-x_0}< \delta$ fulfills 
$\lim\limits_{t\to\infty} \norm {\tilde x_t-x_t} = 0$. $x$ is called 
\emph{asymptotically stable} if it is stable and attractive.

Recall that a \emph{contraction} is a self-mapping on some set with Lipschitz constant
$L<1$, i.e. a mapping $\bar T:M\to M$, $M\subset \R^n$ 
with $\norm{\bar T(x)-\bar T(y)} \leq L \norm {x-y}$ for all 
$x,y\in M$. If $M$ is complete, i.e. all Cauchy sequences converge, then 
a unique fixed point $\xstar\in M$ of $\bar T$ exists by the Banach fixed point theorem. 


\begin{theorem}
\label{th:Eigenvalues-Convergence}
\textbf{Linearized asymptotic stability implies local nonlinear stability}
Consider $\bar T: M \to M$ with a fixed point $\xstar$ and $\bar T$ 
continuously differentiable
in an open neighbourhood $B_r(\xstar) \subset M$ of $\xstar$. 
Denote the Jacobian by $D\bar T_{\xstar}$, and assume 
$\norm {D\bar T_{\xstar}} < 1$ for some norm on $\R^{n\times n}$. 
Then there exists $0 <\varepsilon \leq r$ 
and $0\leq c < 1$ such that for all $x_0$ with $\norm {x_0-\xstar} < \varepsilon$
\[
	\norm {x(t; x_0) - \xstar} \leq c^t \norm {x_0-\xstar} \qquad \forall t\in \N_0. 
\]
i.e. $\xstar$ is locally exponentially and asymptotically stable.
\end{theorem}
The theorem is the core of the first method of Lyapunov for discrete time systems. 
For a proof see \cite[Corollary 4.35.]{Elaydi.2005}.

\section{Convergence Proof}
\label{sec:Convergence-Proof}
Let $\weight \in \R^n$ be the weights of the function $f\left( \weight \right) \in C^2 \left(\R^n, \R \right)$, 
which has to be minimized. We also define $g\left( \weight \right) := \nabla f (\weight) \in \R^n$ as the gradient of $f$ 
and the state variable of our dynamical system 
$x= \left(m,v,w \right)$. With these definitions we can rewrite the ADAM-Optimizer as a system of the form 
\refb{eqDiscreteTimeVariant}.
\begin{align}
\label{eq:OriginalSystemADAM}
m_{t+1} &:= \beta_1 m_t + \left( 1-\beta_1 \right) g \left( \weight_t \right) \nonumber \\
v_{t+1} &:= \beta_2 v_t + \left( 1 - \beta_2 \right) g \left( \weight_t \right)^2\\
\weight_{t+1} &:= \weight_t - \learningrate \frac{\sqrt{1-\beta_2^{t+1}}}{\left( 1-\beta_1^{t+1} \right)} \frac{m_{t+1}}{\sqrt{v_{t+1} + \epsilon^2}}\nonumber
\end{align}
So the ADAM optimizer can be written as the iteration of a time-variant dynamical system
$x_{t+1} = [m_{t+1}, v_{t+1}, \weight_{t+1}]^\transpose  = T \left(t,x \right) = T \left(t,[m_t,v_t, \weight_t]^\transpose \right) \in \R^{3n}$.
We split the system into an autonomous and a non-autonomous part
\begin{align}
\label{eq:System_Origin}
x_{t+1} = T\left(t,x_t \right) = \bar{T} \left( x_t \right) + \Theta \left( t, x_t \right)
\end{align}
with
\begin{align}
\label{eq:AutonomousSystemADAM}
\bar{T} \left( x_t \right) &= \begin{bmatrix}
\beta_1 m_t + \left( 1-\beta_1 \right)g \left( \weight_t \right)\\
\beta_2 v_t + \left( 1-\beta_2 \right)g \left( \weight_t \right)^2\\
\weight_t - \learningrate \frac{m_{t+1}}{\sqrt{v_{t+1}+ \epsilon^2}}
\end{bmatrix}
\end{align}
and
\begin{align}
\label{eq:NonAutonomousSystemADAM}
\Theta \left( t,x_t\right) &= [0,0, \learningrate \Theta_{3} \left( t,x_t\right)]^\transpose\\
\Theta_{3} \left( t,x_t\right) &= \left(1- \frac{\sqrt{1-\beta_2^{t+1}}}{1-\beta_1^{t+1}} \right) \frac{m_{t+1}}{\sqrt{v_{t+1}+ \epsilon^2}} \notag
\end{align}

To avoid lengthy expressions we use $m_{t+1}$ and $v_{t+1}$ as an abbreviation
for the updated terms instead of the filters depending on $m_t$, $g(\weight_t)$ and
$v_t$. The autonomous system is ADAM without bias correction, the disturbance term $\Theta$
adds bias correction which leads to a non-autonomous system.
The Jacobian matrix of the autonomous system \refb{eq:AutonomousSystemADAM} is
\begin{align*}
J_{\bar{T}} \left( m_t, v_t, \weight_t \right) &= \begin{bmatrix}
\beta_1 I & 0 & \left( 1-\beta_1 \right) \nabla_w g \left( \weight_t \right)\\
0 & \beta_2 I & \frac{\partial v_{t+1}}{\partial w_{t}}\\
\frac{\partial w_{t+1}}{\partial m_{t}} & \frac{\partial w_{t+1}}{\partial v_{t}} & \frac{\partial w_{t+1}}{\partial w_{t}}
\end{bmatrix}
\end{align*}
with 
\begin{align*}
\frac{\partial v_{t+1}}{\partial w_t} =& 2 \left( 1-\beta_2 \right) \diag(g\left( \weight_t \right)) \nabla_w g\left( \weight_t \right)\\
\frac{\partial w_{t+1}}{\partial m_{t}} =& -\learningrate \diag \frac{\beta_1}{\sqrt{v_{t+1} + \epsilon^2}}
\\
\frac{\partial w_{t+1}}{\partial v_{t}} =& \frac{\learningrate \beta_2}{2} \diag \left(\frac{m_{t+1}}{\left( v_{t+1} + \epsilon^2\right)^{\frac 3 2}} \right)
\\
\frac{\partial w_{t+1}}{\partial w_{t}} =&  I - \learningrate 
\left(\left(1-\beta_1 \right) 
\diag(v_{t+1} + \epsilon^2)^{-\frac 1 2}   \right. 
\\ 
 &\left. - \diag \left(m_{t+1} (v_{t+1} + \epsilon^2)^{-\frac 3 2} g\left(\weight_t\right) \right) \right) \cdot \nabla_w g\left(\weight_t\right)
\end{align*}
We have the following simple observation:
\begin{lemma}
\label{eq:CriticalMeansFixedPoint}
Consider a critical point $\weightstar$ for $f$, $\nabla f(\weightstar) = 0$. 
Then $\xstar=(0,0,\weightstar)^\transpose$ is a fixed point for \refb{eq:AutonomousSystemADAM}
and \refb{eq:OriginalSystemADAM}. 

Conversely, if  $\xstar=(0,0,\weightstar)^\transpose$ is a fixed point then 
$\nabla f(\weightstar) = 0$.
\end{lemma}
\begin{proof}
We start the iteration with $\weight_0=\weightstar$, 
$v_0 =0$ and $m_0 = 0$, i.e. $x_0=(0,0,\weightstar)$. Then \refb{eq:AutonomousSystemADAM} gives
$x_1 = T(x_0) = x_0$, and inductively $x_t=x_0$ for all $t$. The same holds for
\refb{eq:OriginalSystemADAM}.

Conversely, solving the iteration in $\xstar=(0,0,\weightstar)^\transpose$ for
 $\nabla f(\weightstar)$ immediately gives $\nabla f(\weightstar)=0$.
\end{proof}
Now we investigate the stability of this fixed point with the goal of asymptotic
stability for local minima $\weightstar$. The analysis is simplified because the $m$ and $v$ components of $\xstar$ are 0. So we reach the following Jacobian:
\begin{align*}
J_{\bar{T}} \left( 0,0, \weightstar \right) = \begin{bmatrix}
\beta_1 I & 0 & \left( 1-\beta_1\right) \nabla_w g\left( \weightstar \right)\\
0 & \beta_2 I & 0\\
\frac{- \learningrate \beta_1}{\epsilon} I & 0 & I-\frac{\learningrate \left(1-\beta_1 \right)}{\epsilon} \nabla_w g\left(\weightstar \right)
\end{bmatrix}
\end{align*}
\begin{theorem}
\label{th:Eigenvalues}
Let $J_{\bar{T}} \left( m, v,\weight \right) \in \Mat_{3n}$ be the Jacobian of system \refb{eq:AutonomousSystemADAM} and $\weightstar\in \R^n$ a minimum of $f$
with positive definite Hessian $\nabla^2_w f(\weightstar)=\nabla_w g(\weightstar)$. 
Denote $\eigenvalueshessian_i \in \R$ with $i= 1, \ldots, n$, the $i$-th eigenvalue 
of $\nabla_w g \left( \weightstar \right)$, $\varphi_i = \frac{\learningrate \mu_i}{\epsilon} \left(1-\beta_1\right)$ and all other parameters are defined as in Algorithm \ref{alg:ADAM}. Then $J_{\bar{T}} \left(0,0, \weightstar \right)$ has the eigenvalues, for
$i=1, \ldots, n$:
\begin{align*}
\lambda_{1;i} &= \beta_2\\
\lambda_{2,3;i} &= 
\frac{\left( \beta_1 +1 - \varphi_i \right) \pm \sqrt{\left( \beta_1+1 - \varphi_i \right)^2 - 4 \beta_1}}{2}
\end{align*}
\end{theorem}
In the combination of Theorem \ref{th:Eigenvalues} and \ref{th:Eigenvalues-Convergence} we still have to show, that $|\lambda_{j,i}| < 1$ holds, then the spectral radius for the Jacobian is smaller than $1$ and we prove local convergence.
\begin{theorem}
\label{th:Spectral-radius-Jacobian}
Let the parameters be defined as in Theorem \ref{th:Eigenvalues} and inequality (\ref{eq:Our_Inequality}) holds, then
$
\rho \left( J_{\bar{T}} \left( 0,0,\weightstar \right) \right) < 1.
$
\end{theorem}
\begin{corollary}
\label{cor:Convergence}
Let the parameters be defined as in Theorem \ref{th:Eigenvalues} and such that $\frac{\alpha}{\epsilon} \max_{i=1}^n \left(\eigenvalueshessian_i\right) \left(1-\beta_1 \right)< 2\beta_1 +2$ holds
for $i \in \{1, \ldots, n\}$, then Algorithm \ref{alg:ADAM} converges locally with exponential 
rate of convergence.
\end{corollary}
\begin{proof}
\label{pr:Corollary}
Consider the non-autonomous system \refb{eq:System_Origin} with $\bar{T}(x_t)$ and $\Theta(t,x_t)$ as defined in equations \refb{eq:AutonomousSystemADAM} and \refb{eq:NonAutonomousSystemADAM}. The Hessian of $f$ is continuous, so the gradient
of $f$ is locally Lipschitz with some constant $L>0$, 
$\norm{g(\weight_1) - g(\weight_2)}\leq L\norm {\weight_1-\weight_2}$ for all 
$\weight_1,\weight_2$ in some neighbourhood of $\weight_\star$. Let all other parameters be defined as in Theorem \ref{th:Eigenvalues}, especially $\frac{\learningrate}{\epsilon} \max_{i=1}^n \left(\eigenvalueshessian_i\right) \left( 1-\beta_1\right) <2\beta_1 +2$. Using $m_\star=0$ and $g(\weight_\star) = 0$ we estimate 
\begin{align*}
&\norm{\Theta \left(t,x \right)}\\
&= 
\learningrate \left| \frac{\sqrt{1-\beta_2^{t+1}}}{1-\beta_1^{t+1}} -1 \right| \cdot \frac{\norm{\beta_1 m+(1-\beta_1) g(\weight)}}
{\sqrt{\beta_2 v+(1-\beta_2) g(\weight)^2 +\epsilon^2}}\\
&\leq 
\frac{\learningrate}{\epsilon} 
\left|\frac{ \sqrt{1-\beta_2^{t+1}} - \left( 1-\beta_1^{t+1}\right)}
{\left(1-\beta_1^{t+1}\right)}\right| \cdot \norm{\beta_1 m+(1-\beta_1) g(\weight)}\\
&\leq 
\frac{\learningrate}{\epsilon \left( 1-\beta_1\right)} \left| \frac{\left( 1-\beta_2^{t+1} \right) - \left( 1-\beta_1^{t+1} \right)^2}{\sqrt{1-\beta_2^{t+1}} + 
\left( 1-\beta_1^{t+1}\right)} \right|\\ 
&\quad\cdot\left(\beta_1 \norm{m} + (1-\beta_1)\norm{g(\weight)} \right)\\
&\leq \frac{C}{4} \left| \left( 1-\beta_2^{t+1} \right) - \left( 1-\beta_1^{t+1} \right)^2 \right| \\
&\quad\cdot\left(\beta_1 \norm{m-m_\star}+ (1-\beta_1) \norm{g(\weight)-g\left(\weight_\star\right)} \right)\\
&\leq \frac{C}{4} \left| -\beta_2^{t+1} -2\beta_1^{t+1} + \beta_1^{2(t+1)} \right| \\ 
&\quad\cdot(\beta_1\norm{m-m_\star}+(1-\beta_1) L \norm{\weight-\weight_\star})\\ 
&\leq C\beta^{t+1} \big(\beta_1\norm{m-m_\star} +(1-\beta_1) L \norm{\weight-\weight_\star}\big)
\end{align*}
where we have used the Lipschitz continuity of $g$, and set $\beta = \max \{\beta_1, \beta_2, \beta_1^2\}$, $C := \frac{4 \learningrate}{\epsilon\left(1-\beta_1\right)\left( \sqrt{1-\beta_2} + \left( 1-\beta_1\right) \right)}$. 
The term $\beta_1\norm{m-m_\star}+(1-\beta_1) L \norm{\weight-\weight_\star}$ corresponds to a norm 
$$
\norm {(\tilde m,\tilde\weight)}_{*}:= 
	\beta_1 \norm {\tilde m}
	+ (1-\beta_1)L \norm {\tilde \weight}, \quad \tilde m, \tilde \weight \in \R^n
$$
on $\R^{2n}$ (which does not depend on $\weight_\star$).
By the equivalence of norms in finite dimensional spaces we can 
estimate $\norm {(\tilde m,\tilde\weight)}_{*} 
\leq \tilde C \norm {(\tilde m,\tilde\weight)}$ for some $\tilde C>0$. We continue the 
estimate:
\begin{align*}
\phantom{\norm{\Theta \left(t,x \right)}} 
\leq& C \beta^{t+1}\tilde C \norm{( m-m_\star, \weight-\weight_\star)}
\\
\leq& (C \beta\tilde C) \beta^t \norm{ x-x_\star}
=: \bar C \beta^t \norm{ x-x_\star}
\end{align*}
for some $\bar C>0$. 
With this estimate and Theorem \ref{th:ConvergencePerturbation}, it is sufficient to prove exponential stability
of a fixed point of $\bar{T}$. 
By Theorem \ref{th:Spectral-radius-Jacobian} we get 
$
\rho \left( J_{\bar{T}} \left( 0,0,\weightstar \right) \right) < 1
$. Thus with Theorem \ref{th:Eigenvalues-Convergence} the fixed point 
$(0,0,\weightstar)$ corresponding to the minimum $\weightstar$ is 
locally exponentially stable,
and Theorem \ref{th:ConvergencePerturbation} gives local exponential convergence of
the non-autonomous system $T \left(t,x\right)$, i.e. the ADAM algorithm.
\end{proof}

\subsection{Original ADAM Formulation}
\label{subsec:ConnectionAdam}
We recall, that we use the ADAM optimizer with $\frac{1}{\sqrt{v + \epsilon^2}}$ instead of $\frac{1}{\sqrt{v}+\epsilon}$. Thus line $5$ from Algorithm \ref{alg:ADAM} changes to line $5'$.

\begin{align*}
5': w_{t+1} = w_t - \learningrate \frac{\sqrt{1-\beta_2^{t+1}}}{\left(1-\beta_1^{t+1}\right)} \frac{m_{t+1}}{\sqrt{v_{t+1}} + \epsilon}
\end{align*}

In order to prove the convergence of the original version, reference is made at this point to \cite[Theorem 2.7]{Bof.2018}. Since \cite{Bof.2018} is only available as a non-refereed preprint, the proof can be found in the appendix (Theorem \ref{th:Bof_27xstar}).\\
Consider  the ADAM optimizer without the bias correction  as in system \refb{eq:DiscreteTimeInvariant}, that is with the term $\frac{1}{\sqrt{v+\epsilon^2}}$
in the $\weight$ equation of $x_{t+1} = \bar{T}\left(x_t\right)$  with $x = [m, v, w ]^\transpose$.
From  Corollary \ref{cor:Convergence} we know that $\xstar$ is locally exponentially stable. 
With small changes we can modify this system to the original ADAM formulation 
of \cite{Kingma.2015} without bias correction.
\begin{align}
\label{eq:AdamWithoutBias}
\tilde{x}_{t+1} = \bar{T}(x_t)+ h(x_t) , \quad t \in \N_0, x_0 \in M
\end{align}
with the autonomous system $\bar{T}\left( x_t\right)$ as defined in equation \refb{eq:AutonomousSystemADAM} and 
\begin{align}
\label{eq:DisturbanceToAdam}
h(x_t) &:= \left[0, 0, h_{3}\left(x_t\right)\right]^\transpose\\
h\left(x_t\right) &= \learningrate m_{t+1} \left( \frac{1}{\sqrt{v_{t+1}} + \epsilon} - \frac{1}{\sqrt{v_{t+1}+\epsilon^2}} \right)
\notag
\end{align}
Note that we define the original ADAM without bias correction in dependence of the 
system \refb{eq:DiscreteTimeInvariant} studied so far and as an autonomous system. By means of this consideration, we can prove the local convergence of system \refb{eq:AdamWithoutBias}.
\begin{corollary}
\label{cor:ConvergenceAdamWithoutBias}
Assume the original ADAM without bias correction $\tilde{x}_{t+1} = \bar{T}\left(x_t \right) + h \left(x_t \right)$ as defined in equation \refb{eq:AdamWithoutBias} and $f: \R^n \rightarrow \R$ strictly convex with minimum $w_\star \in \R^n$. Assume $f \in C^2$ and $\nabla^2 f \left( w_\star \right)$ positive definite. Assume inequality \refb{eq:Our_Inequality} holds for the hyperparameters. Then the original ADAM without bias correction converges locally with exponential rate of convergence.
\end{corollary}
\begin{proof}
We start the proof by considering the disturbance $h(x_t)$ from equation \refb{eq:DisturbanceToAdam}. We can estimate the difference in $h(x_t)$ as follows:
\begin{align*}
&\left|\frac{1}{\sqrt{v}+\epsilon} - \frac{1}{\sqrt{v+\epsilon^2}} \right|\\
=& \left|\frac{\sqrt{v+\epsilon^2}-\left(\sqrt{v}+\epsilon\right)}{\sqrt{v+\epsilon^2}\left(\sqrt{v}+\epsilon\right)}\right|\\
=& \left|\frac{\left(\sqrt{v+\epsilon^2}-\sqrt{v}+\epsilon\right)\left(\sqrt{v+\epsilon^2}+\sqrt{v}+\epsilon\right)}{\sqrt{v+\epsilon^2}\left(\sqrt{v}+\epsilon\right)\left(\sqrt{v+\epsilon^2}+\sqrt{v}+\epsilon\right)}\right|\\
=& \frac{2 \sqrt{v} \epsilon}{\sqrt{v+\epsilon^2}\left(\sqrt{v}+\epsilon\right)\left(\sqrt{v+\epsilon^2}+\sqrt{v}+\epsilon\right)}\\
\leq& \frac{2\sqrt{v}\epsilon}{\sqrt{v}\epsilon\left(2\epsilon\right)}
 = \frac{1}{\epsilon}
\end{align*}
The estimate holds
for any $v$, also for $v=0$, and especially for $v_{t+1}$ which appears in $h_3$.
Therefore we can estimate the whole disturbance $\norm{h\left(x_t\right)}$ 
\begin{align*}
&\norm{h\left(x_t\right)} \\
\leq& \frac{\learningrate}{\epsilon} \norm{m_{t+1}-m_\star}\\
=& \frac{\learningrate}{\epsilon} \norm{\beta_1 m_t + \left(1-\beta_1\right) g\left(w_t\right) - m_\star}\\
\leq& \frac{\learningrate}{\epsilon}  \left(\norm{\beta_1 m_t-m_\star} + \left(1-\beta_1\right) \norm{g\left(w_t\right)}\right)\\
\leq& \frac{\learningrate}{\epsilon}  \left( \beta_1 \norm{m_t-m_\star} + \left(1-\beta_1\right) \norm{g\left(w_t\right)-g\left(w_\star\right)}\right)\\
\leq& \frac{\learningrate}{\epsilon}  \left( \beta_1 \norm{m_t-m_\star} + \left(1-\beta_1\right) L \norm{w_t-w_\star}\right)\\
\leq& \frac{\learningrate}{\epsilon}  \beta_1 \norm{m_t-m_\star}\\
\leq& C\left(\epsilon\right) \norm{x_t-\xstar}
\end{align*}
with $C\left(\epsilon\right) = \frac{\learningrate}{\epsilon}$. 
Thus, the first assumption of Theorem \ref{th:ExponentialConvergenceUnderDisturbances} is fulfilled. We know with the Theorems \ref{th:Eigenvalues-Convergence} and \ref{th:Spectral-radius-Jacobian}, that the undisturbed system \refb{eq:AutonomousSystemADAM} converges locally and exponentially to $\xstar$. With this knowledge and the estimation above, we can apply Theorem \ref{th:ExponentialConvergenceUnderDisturbances}. It follows that $\tilde{x}_\star = \xstar$ is a global exponentially stable fixed point for the system \refb{eq:AdamWithoutBias}. \end{proof}

Involving the bias correction, we have to adjust system \refb{eq:AdamWithoutBias} to
\begin{align}
\label{eq:OriginalAdam}
\tilde{x}_{t+1} = T\left(x_t\right) + h\left(x_t \right) + \tilde{\Theta} \left(t,x\right)
\end{align}
with
\begin{align}
\label{eq:BarTheta}
\tilde{\Theta} \left(t,x\right) &:= \left[ 0,0, \learningrate \tilde{\Theta}_{3} \left(t,x\right)\right]^\transpose\\
\tilde{\Theta}_{3} \left(t,x\right) &= \left(1-\frac{\sqrt{1-\beta_2^{t+1}}}{1-\beta_1^{t+1}} \right) \frac{m_{t+1}}{\sqrt{v_{t+1}} + \epsilon} \notag
\end{align}

\begin{corollary}
Assume the original ADAM with bias correction as the non-autonomous system $\tilde{x}_{t+1} = \bar{T} \left(x_t \right) + h \left(x_t \right) + \tilde{\Theta} \left(t,x\right)$ like defined in equation \refb{eq:OriginalAdam}. All other assumptions are adopted by Corollary \ref{cor:ConvergenceAdamWithoutBias}. Then system \refb{eq:OriginalAdam} converges locally with exponential rate of convergence.
\end{corollary}
\begin{proof}
Assume the non-autonomous system \refb{eq:NonAutonomousSystemADAM} with $\bar{T}\left(x_t\right)$, $h\left(x_t\right)$ and $\tilde{\Theta}\left(t,x\right)$ as defined in equations \refb{eq:AutonomousSystemADAM}, \refb{eq:DisturbanceToAdam} and \refb{eq:BarTheta}. Since the estimation of $\norm{\tilde{\Theta}\left(t,x\right)}$ is analogous to the estimation in the proof of Corollary \ref{cor:Convergence}, reference is made here to this proof and the estimation of $\norm{\tilde{\Theta}\left(t,x\right)}$ is given by
\begin{align*}
\norm{\tilde{\Theta}\left(t,x\right)} \leq& C \beta^{t+1} (\beta_1 \norm{m-m_\star} \\
&+ \left(1-\beta_1 \right) L \norm{w - w_\star})
\end{align*}
Here $L > 0 $ is the Lipschitz constant of $f$, $\beta = \max \{\beta_1, \beta_2, \beta_1^2 \}$ and $C:= \frac{4 \learningrate}{\epsilon \left(1-\beta_1\right) \left(\sqrt{1-\beta_2}+\left(1-\beta_1\right)\right)}$. Analogous to the proof of Corollary \ref{cor:Convergence} we can shorten the inequality by the equivalence of norms in finite dimensional spaces. Thus 
\begin{align*}
\norm{\tilde{\Theta}\left(t,x\right)} \leq \bar{C} \beta^t \norm{x-\xstar}
\end{align*}
for some $\bar{C} > 0$.
With Corollary \ref{cor:ConvergenceAdamWithoutBias} we get the exponential stability of a fixed point of $\tilde{x}_{t+1} = \bar{T}\left(x_t\right) + h \left(x_t\right)$. Combining this result with the upper estimate and Theorem \ref{th:ConvergencePerturbation}, we get local exponential convergence of the non-autonomous system \refb{eq:OriginalAdam}, i.e. the ADAM algorithm with bias correction defined as in \cite{Kingma.2015}.
\end{proof}
\section{Extension to other optimizers}
\label{sec:Extension other optimiziers}
In the category of adaptive gradient descent algorithms are a several popular ones, 
where ADAM is only one of them. Some of them can also be seen as momentum methods, and they are all available in a stochastic or mini-batch way. We will also apply our convergence proof to some of these algorithms in the complete batch mode. This shows how general the methodology of the proof is and shows that other algorithms can also benefit from it.
To keep this work clear, we note these algorithms in the elegant way from \cite{Reddi.2019}. We assume the Generic Adaptive Method Setup from Algorithm \ref{alg:GAMS} and set only $v_t$ and $m_t$ different for each optimizer (see Table \ref{tab:optimizer extension}). Note that we implemented a function $d(m_{t+1})$, which is only interesting for AdaDelta. This is necessary due to the use of $m_t$ in the weight update of $w_{t+1}$.
It will become clear that the essential requirement of our method are stable eigenvalue 
which are present in most but not all algorithms.

\begin{algorithm}
\caption{Generic Adaptive Method Setup}
\label{alg:GAMS}
\begin{algorithmic}[1]
\REQUIRE $\learningrate \in \R^+$, $\epsilon \in \R$, $\weight_0 \in \R^n$ and the function $f(\weight) \in C^2 \left( \R^n, \R \right)$
\STATE $m_0 = 0$, $v_0 = 0$, $t = 0$
\WHILE{$\weight$ not converged}
\STATE $v_{t+1} = \psi \left(v_t, w_t\right) $
\STATE $m_{t+1} = \varphi \left(m_t, w_t\right) $ 
\STATE $\weight_{t+1} = \weight_t - \learningrate \frac{d(m_{t},w_t)}{\sqrt{v_{t+1} + \epsilon^2}}$
\STATE $t = t+1$
\ENDWHILE
\end{algorithmic}
\end{algorithm}

\begin{table*}[ht]
\centering
\caption{Popular adaptive gradient descent optimizers}
\label{tab:optimizer extension}
\begin{tabular}{|C{3.1cm}||c|c|c|c|}
\hline
Algorithm & $\varphi \left(m_t, w_t\right)$ & $\psi \left(v_t, w_t\right)$ & Momentum & d($m_{t}, w_t$)\\
\hline \hline
SGD \cite{Robbins.1951} & $g\left(w_t\right)$ & $1-\epsilon^2$ & $\beta = 1$ & $\varphi \left(m_t, w_t\right) $\\
\hline
RMSProp \cite{Hinton.2012} & $g\left(w_t\right)$ & $\beta v_t + \left(1-\beta \right) g \left(w_t \right)^2$ & $0<\beta<1$ & $\varphi \left(m_t, w_t\right) $\\
\hline
AdaGrad \cite{Duchi.2011} & $g\left(w_t\right)$ & $v_t+g\left(w_t\right)^2$ & $\beta = 1$ & $\varphi \left(m_t, w_t\right) $\\
\hline
AdaDelta \cite{Zeiler.2012} & $\beta m_{t} + \left(1-\beta\right) \cdot $ & $\beta v_{t} + \left(1-\beta \right) g \left(w_t \right)^2$ & $0<\beta<1$ & $g(w_t) \sqrt{m_t + \epsilon^2}$\\
& $g \left(w_t\right)^2 \frac{m_{t}+\epsilon^2}{v_{t+1} +\epsilon^2}$ & & & \\
\hline 
ADAM without  & $\beta_1 m_t + $ & $\beta_2 v_t + \left(1-\beta_2 \right) g \left(w_t \right)^2$ & $0 < \beta_1,\beta_2 < 1$ & $\varphi \left(m_t, w_t\right) $\\
bias-correction \cite{Kingma.2015} & $\left(1-\beta_1 \right) g \left(w_t \right)$ & & & \\
\hline
\end{tabular}
\end{table*}

\subsection*{RMSProp}
First we observe the RMSProp algorithm by Hinton et. al. \cite{Hinton.2012}. We can define it with the following system
\begin{align*}
x_{t+1} = T \left( x_t \right) &= \begin{bmatrix}
\beta v_t + \left( 1-\beta \right)g \left( \weight_t \right)^2 \\
\weight_t - \learningrate \frac{g\left(\weight_t\right)}{\sqrt{v_{t+1}+ \epsilon^2}}
\end{bmatrix}
\end{align*}
Herewith we can calculate the Jacobian analogous to the convergence proof of the ADAM Optimizer.
\begin{align*}
J_{\bar{T}} \left( v_t, \weight_t \right) &= \begin{bmatrix}
\beta_2 I & \frac{\partial v_{t+1}}{\partial w_{t}}\\
\frac{\partial w_{t+1}}{\partial v_{t}} & \frac{\partial w_{t+1}}{\partial w_{t}}
\end{bmatrix}
\end{align*}
with 
\begin{align*}
\frac{\partial v_{t+1}}{\partial w_{t}} =& 2 \left( 1-\beta \right) \diag(g\left( \weight_t \right)) \nabla_w g\left( \weight_t \right)\\
\frac{\partial w_{t+1}}{\partial v_{t}} =& \frac{\learningrate \beta}{2} \diag \left( \frac{g\left(w_t\right) }{ \left( v_{t+1} + \epsilon^2\right)^{\frac 3 2}}\right)
\\
\frac{\partial w_{t+1}}{\partial w_{t}} =&  I - \learningrate 
\left(\diag(v_{t+1} + \epsilon^2)^{-\frac 1 2}   \right. 
\\ 
 &\left. - \diag \left(g\left(w_t\right)(v_{t+1} + \epsilon^2)^{-\frac 3 2} g\left(\weight_t\right) \right) \right) \\
&\cdot \nabla_w g\left(\weight_t\right) 
\end{align*}
This is simplified in the minimum $\weightstar$ to
\begin{align*}
J_{T} \left( 0, \weightstar \right) = \begin{bmatrix}
\beta I & 0\\
0 & I-\frac{\learningrate }{\epsilon} \nabla_w g\left(\weightstar \right)
\end{bmatrix}
\end{align*}
As you can see we reach $n$ times the eigenvalue $\lambda_1 = \beta$, which is per definition smaller than $1$. The submatrix $I-\frac{\learningrate }{\epsilon} \nabla_w g\left(\weightstar \right)$ is again symmetric since $\nabla_w g\left(\weightstar\right)$ is the Hessian of $f$. Therefore we can diagonalize the matrix and get $n$ times the eigenvalue $\lambda_2 = 1-\frac{\learningrate}{\epsilon} \mu_i$. Next step is to analyze the absolute value of the eigenvalues.
\begin{align*}
&&|\lambda_2| &= |1- \frac{\learningrate}{\epsilon} \mu_i|  < 1\\
& \Leftrightarrow & -2 &< -\frac{\learningrate}{\epsilon} \mu_i < 0\\
& \Leftrightarrow & 0 &< \mu_i < \frac{2\epsilon}{\learningrate}
\end{align*}
Due to the fact that we look at eigenvalues of a positive definite matrix, $0<\mu_i$ is clear. So for local convergence with exponential rate in the RMSProp algorithm we only have to fulfill:
\begin{align*}
\max\limits_{i=1}^n (\mu_i) < \frac{2 \epsilon}{\learningrate}
\end{align*}

\subsection*{AdaGrad}
For AdaGrad, a proof in this form is not possible. The reason for this is $\beta=1$. With this missing parameter, the following Jacobian is created.
\begin{align*}
J_{T} \left( v_t, \weight_t \right) = \begin{bmatrix}
I & 2 g(\weight_t) \nabla_w g(\weight_t)\\
\frac{1}{2} \diag\left( \frac{g\left(\weight_t\right)}{\left( v_{t+1} + \epsilon^2 \right)^{\frac{3}{2}}} \right) & I-\frac{\learningrate }{\epsilon} \nabla_w g\left(\weight_t \right)
\end{bmatrix}
\end{align*}
In the minimum $(0,\weightstar)$ 
\begin{align*}
J_{T} \left( 0, \weightstar \right) = \begin{bmatrix}
I & 0\\
0 & I-\frac{\learningrate }{\epsilon} \nabla_w g\left(\weightstar \right)
\end{bmatrix}
\end{align*}
we see that $\lambda_1 = 1$ no matter which $\max\limits_{i=1}^n \left( \mu_i \right)$ we calculate. Thus, no convergence statement is possible using our approach.
\subsection*{AdaDelta}
Finally, we look at the AdaDelta algorithm from \cite{Zeiler.2012}. This algorithm is somewhat more difficult to bring into the system structure. But with different time steps $t$ and $t+1$ for $v$ and $m$ we can write:
\begin{align*}
x_{t+1} = T \left(x_t \right) = \begin{bmatrix}
\beta v_{t} + \left( 1-\beta \right) g\left(w_t\right)^2\\
\beta m_{t} + \left( 1-\beta \right) g\left(w_t\right)^2 \frac{m_{t}+\epsilon^2}{v_{t+1} + \epsilon^2}\\
w_t - \learningrate \frac{\sqrt{m_{t}+\epsilon^2}}{\sqrt{v_{t+1} + \epsilon^2}} g\left(w_t\right)
\end{bmatrix}
\end{align*}
Note that we add the learning rate $\learningrate$ to the optimizer different to the original paper. If we set $\learningrate = 1$ we reach the original formulation from Zeiler \cite{Zeiler.2012}. We will discuss some different learning rates in Section \ref{sec:Experiments}. For $T(x_t)$ we get the  Jacobian
\begin{align*}
J_{T} \left( v_t, m_t, \weight_t \right) &= \begin{bmatrix}
\beta I & 0 & \frac{\partial v_{t+1}}{\partial w_t}\\
\frac{\partial m_{t+1}}{\partial v_t} & \frac{\partial m_{t+1}}{\partial m_t}& \frac{\partial m_{t+1}}{\partial w_t}\\
\frac{\partial w_{t+1}}{\partial v_t} & \frac{\partial w_{t+1}}{\partial m_t} & \frac{\partial w_{t+1}}{\partial w_t}
\end{bmatrix}
\end{align*}
with 

\begin{align*}
\frac{\partial m_{t+1}}{\partial v_{t}} =& - \beta \left(1-\beta\right) \diag \left(
 \frac{g\left(w_t\right)^2 \left( m_{t}+\epsilon^2 \right)}{\left(v_{t+1}+\epsilon^2\right)^2}\right) \\
\frac{\partial m_{t+1}}{\partial m_t} =& \beta I + \left(1-\beta\right) \diag \left( \frac{g\left(w_t\right)^2}{v_{t+1}+\epsilon^2}\right)\\
\frac{\partial m_{t+1}}{\partial w_t} =& 2\left(1-\beta\right) \diag\left(\frac{\left( m_t+\epsilon^2 \right) \left(\beta v_t + \epsilon^2\right) g\left(w_t\right)}{\left(v_{t+1}+\epsilon^2\right)^2}\right)\\
& \cdot \nabla g\left(w_t\right)\\
\frac{\partial v_{t+1}}{\partial w_t} =& 2 \left( 1-\beta \right) \diag(g\left( \weight_t \right)) \nabla_w g\left( \weight_t \right)\\
\frac{\partial w_{t+1}}{\partial v_t} =& \frac{\beta \learningrate}{2} \diag \left( \frac{\sqrt{m_{t}+\epsilon^2} g\left(w_t\right)}{\left(v_{t+1}+\epsilon^2\right)^{\frac{3}{2}}}\right)\\
\frac{\partial w_{t+1}}{\partial m_t} =& -\frac{\learningrate}{2} \diag \left(\frac{g\left(w_t\right)}{\sqrt{v_{t+1}+\epsilon^2}} \frac{1}{\sqrt{m_{t}+\epsilon^2}}\right) \\
\frac{\partial w_{t+1}}{\partial w_t} =&  I - \learningrate \diag \left( \frac{\sqrt{m_{t}+\epsilon^2} }{\sqrt{v_{t+1}+\epsilon^2}} \right.\\
& \left. - \frac{\sqrt{m_{t}+\epsilon^2} \left(1-\beta\right) g\left(w_t\right)^2}{\left( v_{t+1} +\epsilon^2\right)^{\frac{3}{2}}} \right) \nabla_w g\left(w_t \right)
\end{align*}

Fortunately, by inserting $\xstar$, this Jacobian is greatly simplified to
\begin{align*}
J_T \left(0,0,\weight_\star \right) = \begin{bmatrix}
\beta I & 0 & 0\\
0 & \beta I & 0\\
0 & 0 & I- \learningrate \nabla_w g \left(\weight_\star\right)
\end{bmatrix}
\end{align*}
We can easily identify $2n$ times the eigenvalue $\lambda_1 = \beta$ of the Jacobian. We diagonalize the Hessian of $f$ and identify $\lambda_{2,i} = 1- \learningrate \mu_i$. By observing the spectral radius of the Jacobian we see, that $\lambda_1 = \beta$ is by definition between $0$ and $1$. For $\lambda_2$ we generate the following general inequality.
\begin{align*}
0 < \max_{i=1}^n \left(\mu_i \right) < \frac{2}{\learningrate}
\end{align*}
Thus, when the inequality above is satisfied, AdaDelta is locally convergent with exponential rate of convergence.
\subsection*{Conclusion}
These proofs of convergence, show the generality of this method. Only for AdaGrad we are not able to proof the convergence due to the missing decay rate $\beta$. The result can also be used to adjust the hyperparameters for future optimizations. Table \ref{tab:optimizer extension hyperparameter bounding} shows the resulting hyperparameter bounding for each algorithm.

It is also quite astonishing that the convergence behaviour of AdaDelta is not dependent on the parameter $\epsilon$. The convergence behaviour from the relatively similar algorithms ADAM and RMSProp depend heavily on the ratio between $\learningrate$ and $\epsilon$. Many algorithms reduce the learning rate $\learningrate$ over time. For ADAM or RMSProp, however, it might also be useful to increase $\epsilon$. At this point it should be noted that due to the addition with $v$ it is not mathematically equivalent. 
\begin{table*}[ht]
\centering
\caption{Hyperparameter bounds for popular adaptive gradient descent optimizers}
\label{tab:optimizer extension hyperparameter bounding}
\begin{tabular}{|c|c|}
\hline
Algorithm & Hyperparameter bounding\\
\hline \hline
SGD \cite{Robbins.1951} & $\max\limits_{i=1}^n (\mu_i) < \frac{2}{\learningrate}$\\
\hline
RMSProp \cite{Hinton.2012} & $\max\limits_{i=1}^n (\mu_i) < \frac{2 \epsilon}{\learningrate}$\\
\hline
AdaGrad \cite{Duchi.2011} & No convergence statement possible with this method\\
\hline
AdaDelta \cite{Zeiler.2012} & $\max\limits_{i=1}^n (\mu_i) < \frac{2}{\learningrate}$\\
\hline
ADAM without bias-correction \cite{Kingma.2015} & $\frac{\alpha}{\epsilon} \max\limits_{i=1}^n \left(\eigenvalueshessian_i\right) \left(1-\beta_1 \right)< 2\beta_1 +2$\\
\hline
\end{tabular}
\end{table*}
 
\section{Experiments}
\label{sec:Experiments}
For the sake of clarity, in the next two subsection we will look at the ADAM only. We will then apply similar experiments to the other optimizers, which can be found in subsection \ref{subsec:Experiments other optimizers} and in the appendix.
\subsection{Numerical Convergence}
\begin{figure}[!t]
\center
\includegraphics[width=\bildgroesse , trim = 100 250 100 250, clip]{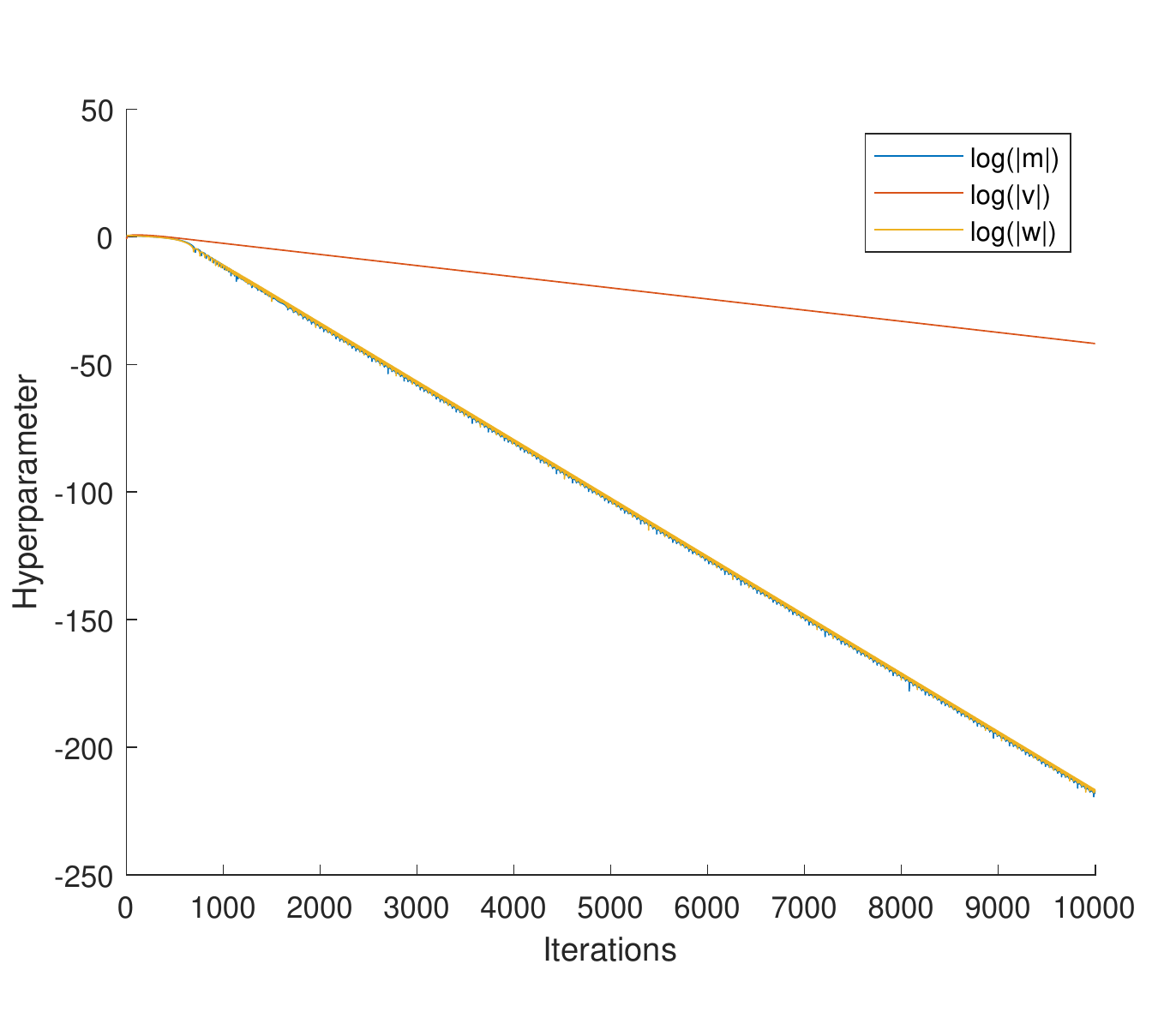}
\caption{Exponential convergence behaviour}
\label{fig:Exponential-Convergnece-Behaviour}
\end{figure}

\begin{figure}[!t]
\center
\includegraphics[width=\bildgroesse , trim = 100 250 100 250, clip]{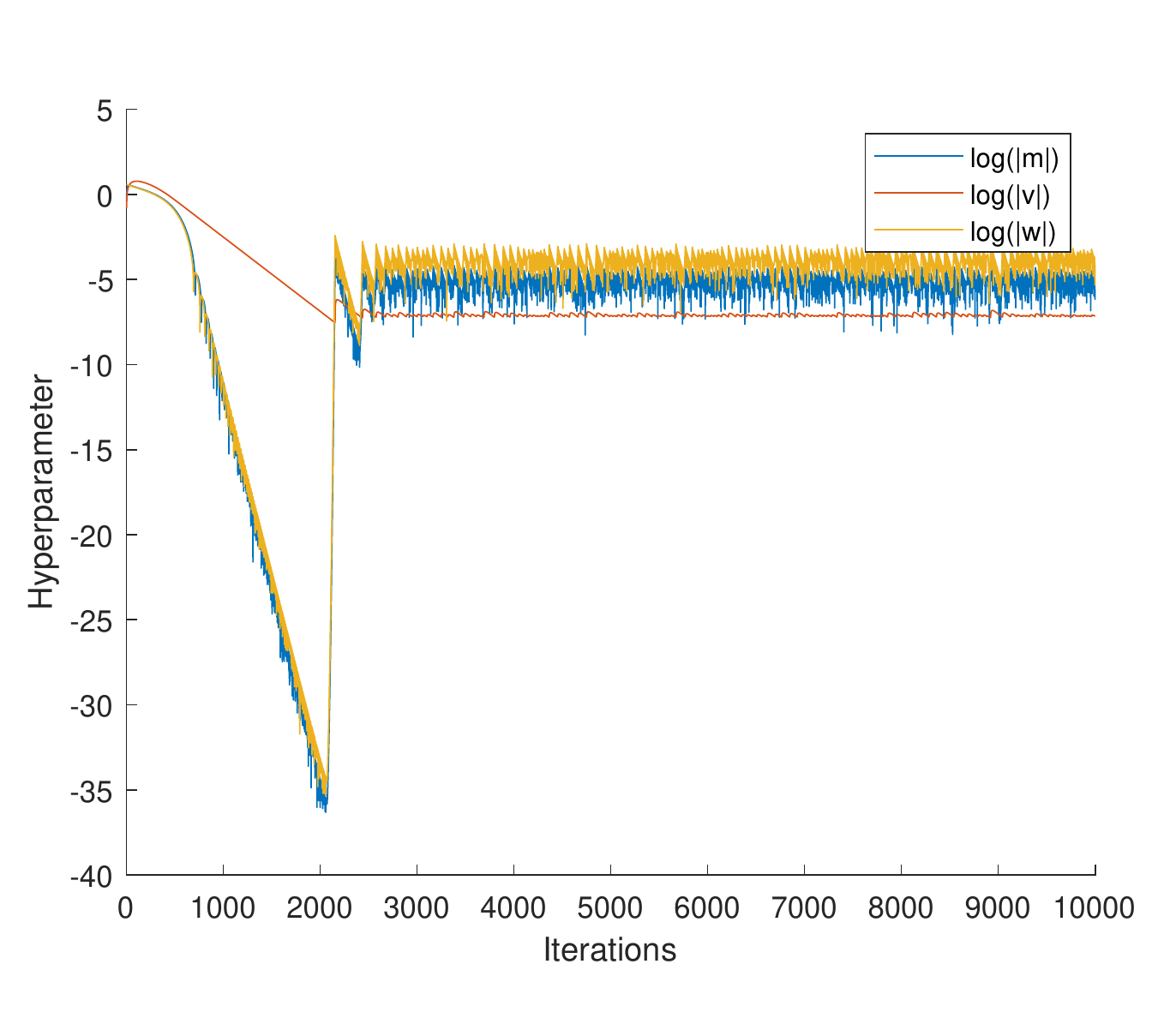}
\caption{Chaotic behaviour}
\label{fig:Chaotic-Behaviour}
\end{figure}

\begin{figure}[!t]
\center
\includegraphics[width=\bildgroesse , trim = 100 250 100 250, clip]{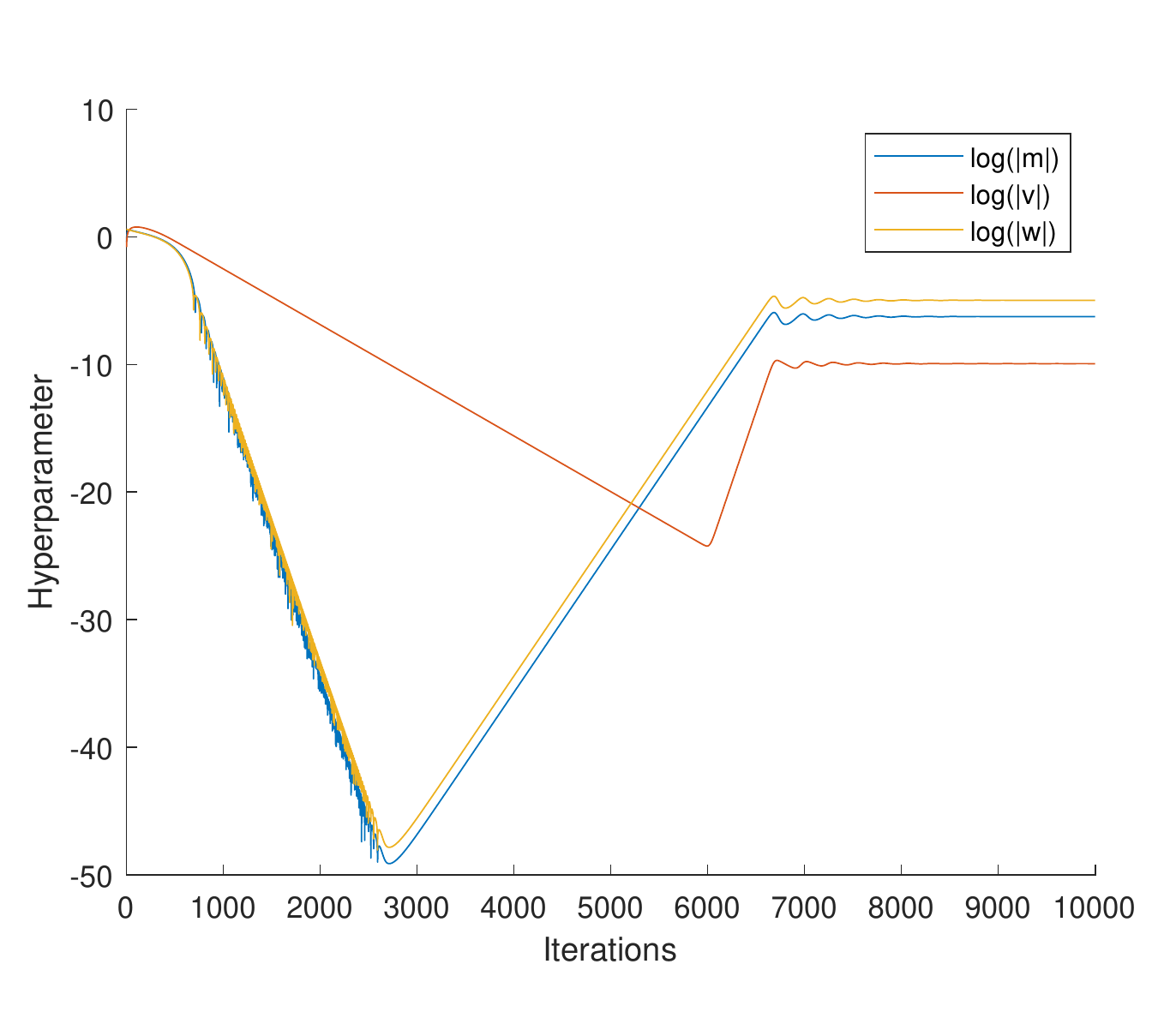}
\caption{Chaotic behaviour near boundary}
\label{fig:Chaotic-Behaviour-near-boundary}
\end{figure}

\begin{figure}[!t]
\center
\includegraphics[width=\bildgroesse]{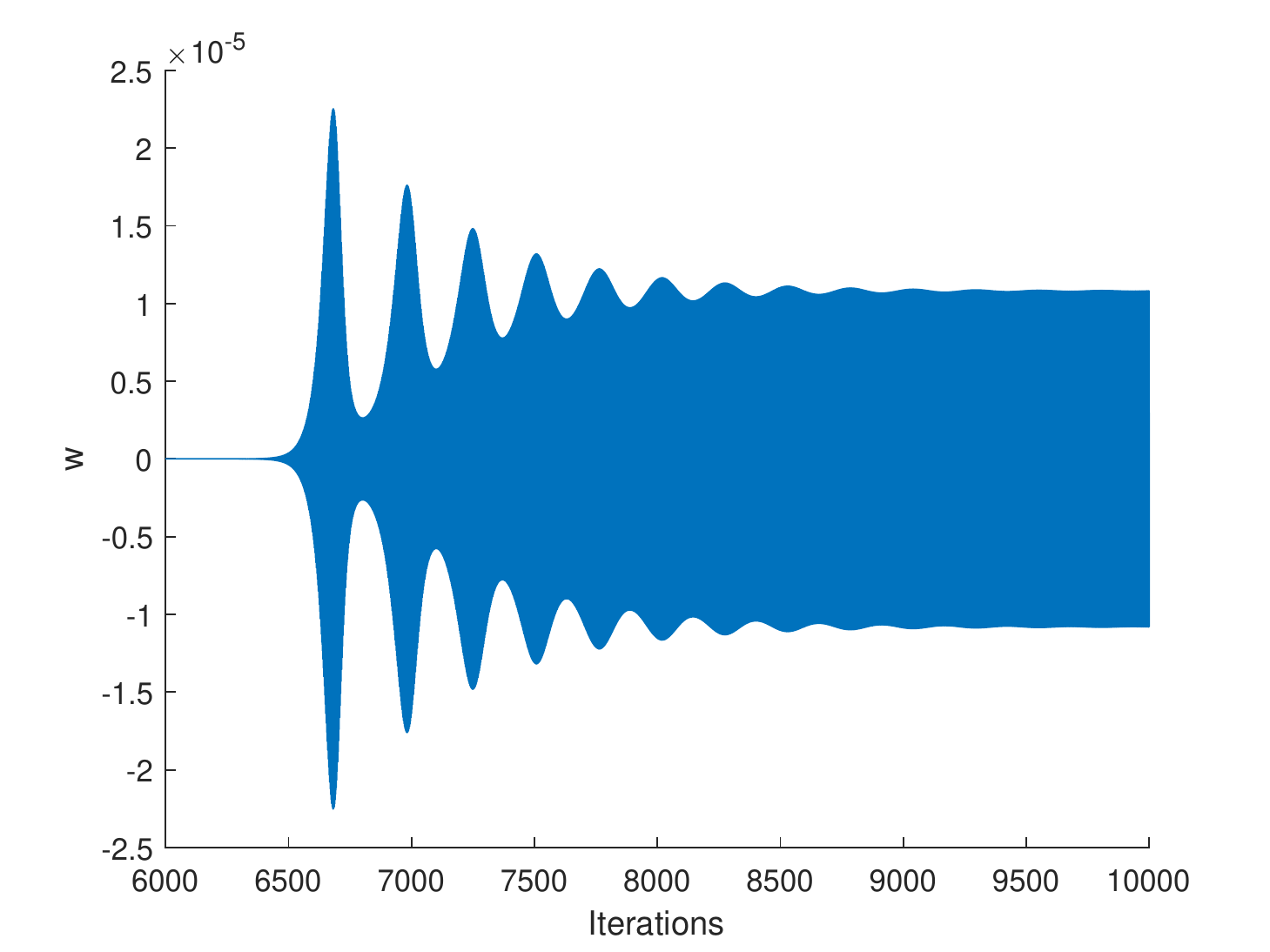}
\caption{Chaotic behaviour of $w$ near boundary}
\label{fig:Chaotic-Behaviour-near-boundary-w}
\end{figure}

The convergence proof in Section \ref{sec:Convergence-Proof} only shows the local convergence under the hyperparameter bound \refb{eq:Our_Inequality}. Whether the boundary is strict or whether there are elements outside this boundary that also converge was not answered.
To study the numerical behaviour of ADAM, we choose $f(x) = \frac{x^2}{2}+10, \R \to \R$ with the hyperparameters $\beta_1 = 0.9, \beta_2 = 0.99, \learningrate = 0.01, m_0 = 0, v_0 = 0$ and $w_0 = 4$. With $\epsilon = 10^{-2}$ inequality \refb{eq:Our_Inequality} holds and ADAM shows exponential convergence behaviour (see Figure \ref{fig:Exponential-Convergnece-Behaviour}).
By choosing $\epsilon = 10^{-8}$ which clearly breaks inequality \refb{eq:Our_Inequality}, we see oscillating behaviour as mentioned in \cite{BockWeiss.2019c} (see Figure \ref{fig:Chaotic-Behaviour}).
If we solve inequality \refb{eq:Our_Inequality} with the assumed hyperparameters to $\epsilon$, we reach the bound $\epsilon < 2.63158 \cdot 10^{-4}$. When slowly reducing $\epsilon$, one can recognize the evolution of chaotic behaviour. Already at $\epsilon = 2.62936 \cdot 10^{-4}$ the exponential convergence is disturbed (see Figure \ref{fig:Chaotic-Behaviour-near-boundary}) and $w$ starts to jump around $w_{\star}$ (see Figure \ref{fig:Chaotic-Behaviour-near-boundary-w}).
We could not find an $\epsilon$ with which our variant converges and the original variant of \cite{Kingma.2015} does not converge or vice versa.
\subsection{Solution Behaviour}
\label{subsec:Solution Behaviour}
To compare our requirements for convergence to the requirements taken by \cite{Reddi.2019} or \cite{Kingma.2015}, we make some empirical experiments. First, we look at the different requirements to the hyperparameter.
\begin{align}
\label{eq:Reddi_Requirement}
\beta_1 &< \sqrt{\beta_2}\\
\label{eq:Kingma_Requirement}
\beta_1^2 &< \sqrt{\beta_2}
\end{align}
Inequality \refb{eq:Our_Inequality} describes the needed requirement presented in this paper. Problematically in this estimation is, that we need the maximum eigenvalue of $\left( 1-\beta_1 \right) \nabla_w g \left( \weightstar \right)$ and consequently $\weightstar$. Therefore our estimate is an a posteriori estimate. But with \refb{eq:Our_Inequality} we learn something about the relationship between the hyperparameters. $\frac{\learningrate}{\epsilon}$ has to be very small to fulfill inequality \refb{eq:Our_Inequality}. With $\learningrate$ small or $\epsilon$  big we always make the weight change smaller and so we do not jump over $\weightstar$. Inequality \refb{eq:Reddi_Requirement} was presented in \cite{Reddi.2019} and inequality \refb{eq:Kingma_Requirement} was originally presented in \cite{Kingma.2015}. Both are a priori estimations for the hyperparameters.

To show the behaviour of all estimations we set up the following experiments. In Experiment $1$ and $2$ we want to minimize $f(\weight) := \weight^4+\weight^3$ with the minimum $\weightstar = -\frac{3}{4}$. In Experiment $3$ we minimize the multidimensional function $f(\weight_1,\weight_2) := \left(\weight_1+2 \right)^2 \left( \weight_2+1 \right)^2 + \left( \weight_1+2 \right)^2 + 0.1 \left( \weight_2+1 \right)^2$ with the minimum $\weightstar = \left(-2,-1 \right)$. We run the ADAM optimizer $10 000$ times in every hyperparameter setting and if the last five iterations $\weight_{end} \in \R^5$ are near enough to the known solution $\weightstar$ the attempt is declared as convergent. Near enough in this setting means that all components of $\weight_{end}$  are contained in the interval $[\weightstar - 10^{-2}, \weightstar + 10^{-2}]$. The color coding of our experiments can be found in Table \ref{tab:Colortable}. To keep the clarity of our results we only compare the original ADAM inequality with our inequality. With inequality \refb{eq:Reddi_Requirement} we obtain similar figures.
\begin{table}
\caption{Co lour description for the convergence investigations}
\label{tab:Colortable}
\center
\begin{tabular}{|c||C{1.65cm}|C{1.65cm}|C{1.65cm}|}
\hline
& Inequality \refb{eq:Kingma_Requirement} satisfied & Inequality \refb{eq:Our_Inequality} satisfied & ADAM finds solution\\
\hline \hline
green & yes & yes & yes \\
blue & no & yes & yes\\
yellow & yes & no & yes\\
white & no & no & yes\\
black & yes & yes & no\\
cyan & no & yes & no\\
magenta & yes & no & no\\
red & no & no &no \\
\hline
\end{tabular}
\end{table}
\subsection*{Experiment 1}
\label{subsec:Experimetn_1}
\begin{figure}[!t]
\center
\includegraphics[width=\bildgroesse , trim = 100 250 100 250, clip]{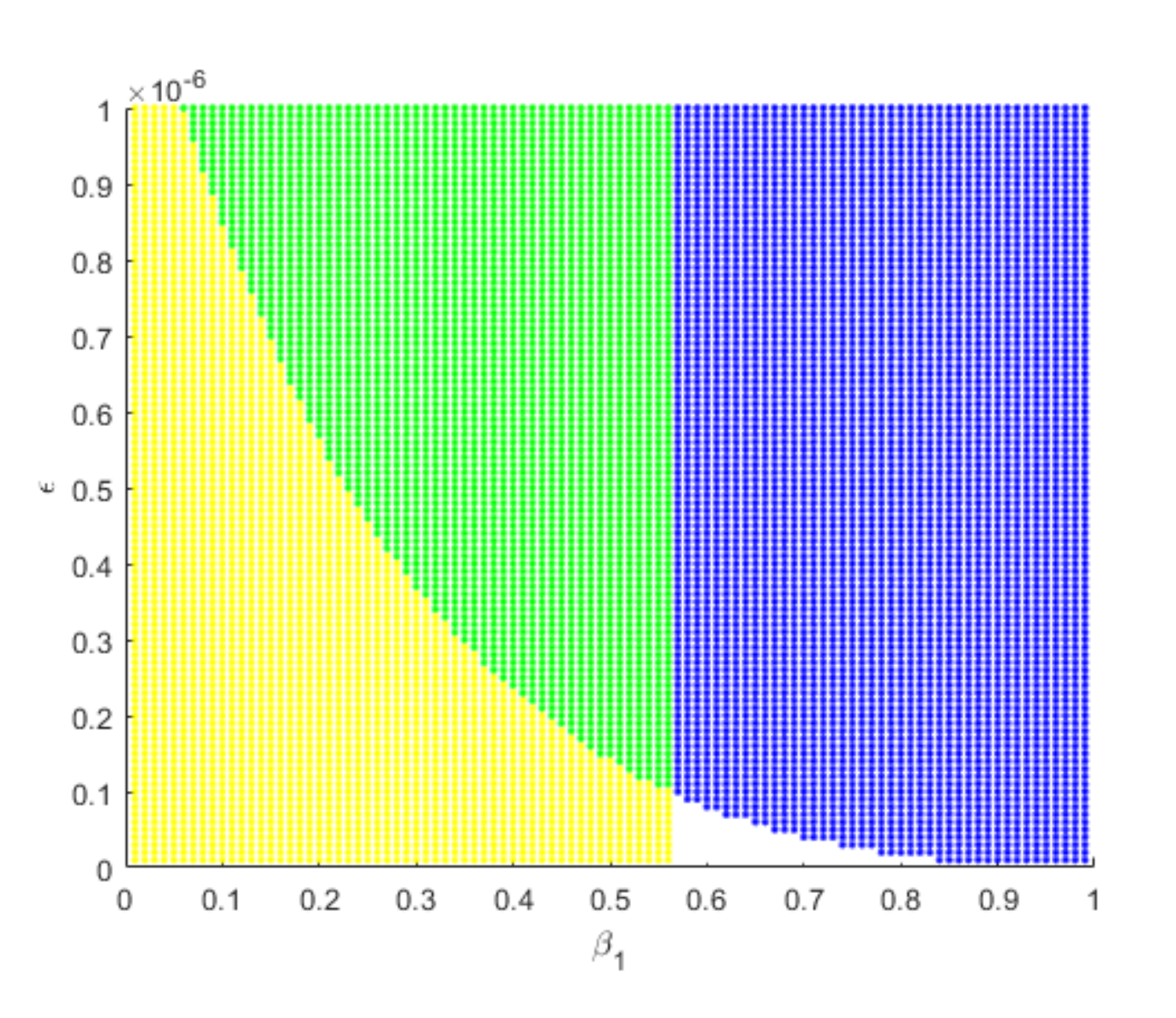}
\caption{Iterating over $\epsilon$ and $\beta_1$}
\label{fig:Experiment1}
\end{figure}

First, we iterate over $\epsilon \in \left\{10^{-4}, \ldots , 10^{-3}\right\}$ and $\beta_1 \in \{0.01, \ldots, 0.99\}$. The other hyperparameters are fixed $\learningrate = 0.001$, $\beta_2 = 0.1$. This setting leads us to figure \ref{fig:Experiment1}.
The only area, where the ADAM optimizer is not finding a solution (red dots), is inside the white area. So both inequalities are not satisfied and the convergence is not given. The white area -- ADAM converge but no inequality is satisfied -- is formed because we only talk about estimation and not clear boundaries. The blue and yellow area can be made larger or smaller by changing $\beta_2$ or $\learningrate$.
\subsection*{Experiment 2}
\label{subsec:Experimetn_2}
\begin{figure}[!t]
\center
\includegraphics[width=\bildgroesse , trim = 100 250 100 250, clip]{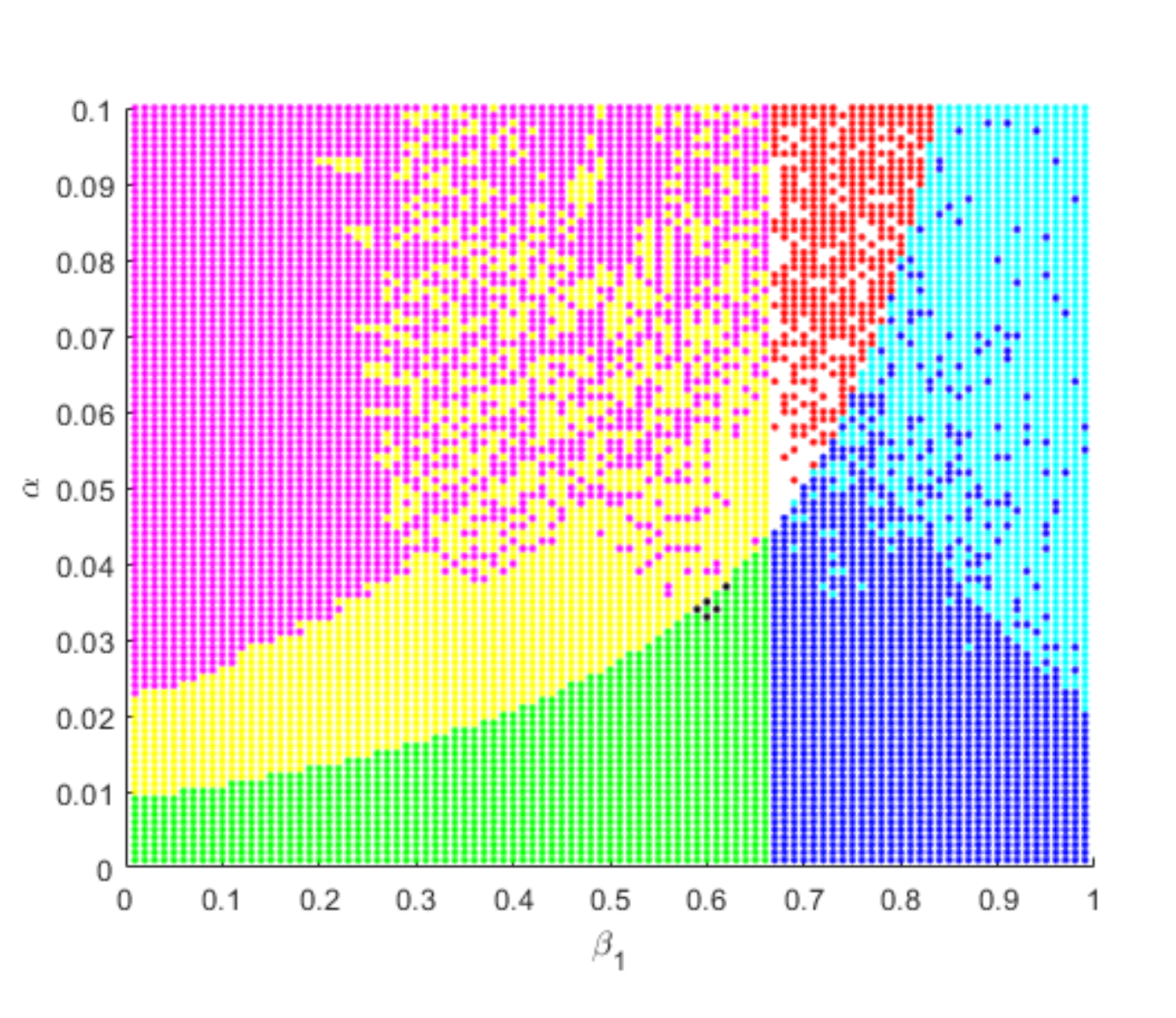}
\caption{Iterating over $\learningrate$ and $\beta_1$}
\label{fig:Experiment2}
\end{figure}

\begin{figure}[!t]
\center
\includegraphics[width=\bildgroesse , trim = 100 250 100 250, clip]{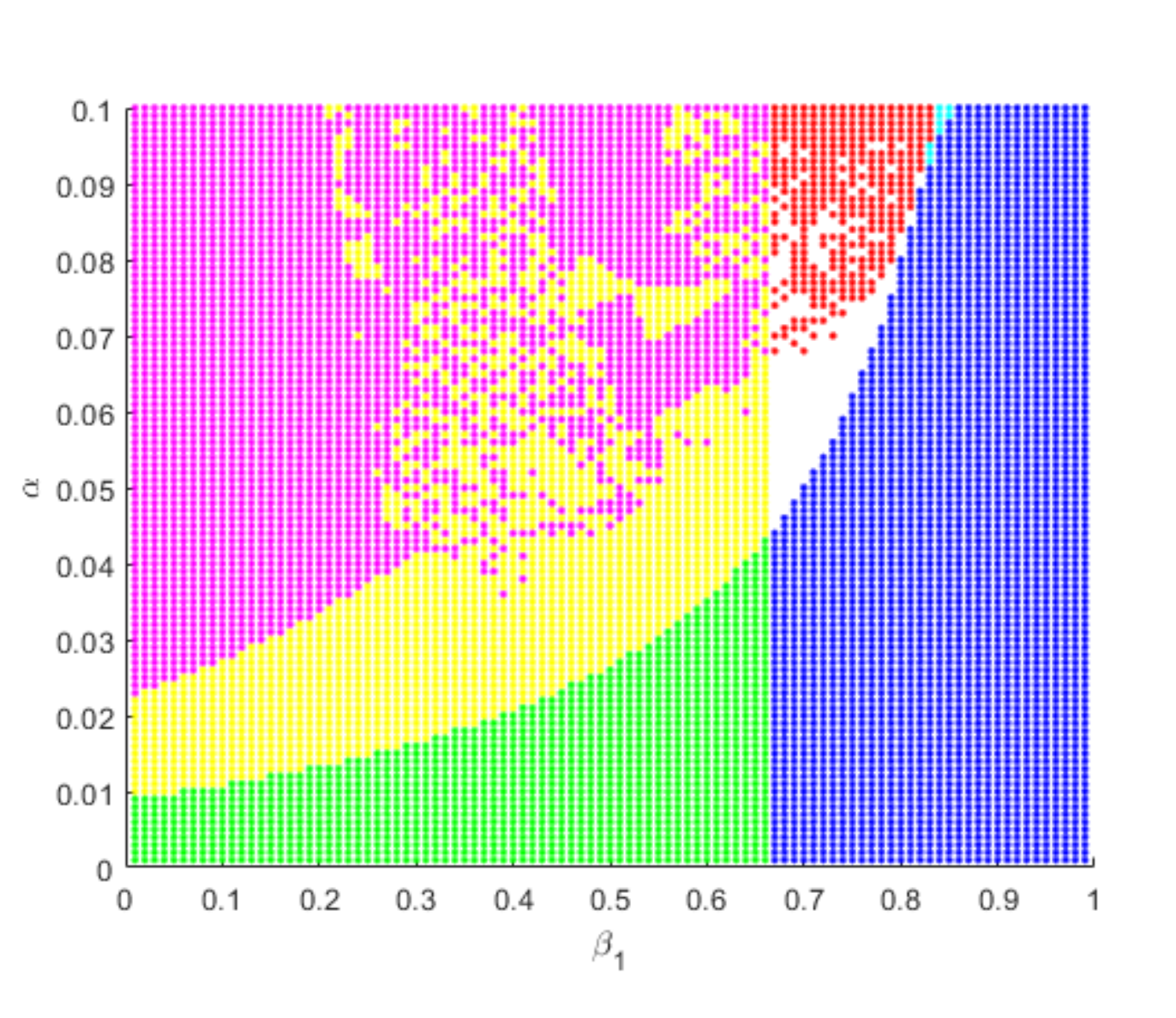}
\caption{Iterating over $\learningrate$ and $\beta_1$ with $x_0 = -0.750000001$}
\label{fig:Experiment2_startvalue_near}
\end{figure}

In the second experiment we iterate over $\learningrate \in \{0.001,\ldots ,0.1\}$ and $\beta_1 \in \{0.01,\ldots ,0.99\}$. $\beta_2 = 0.2$ and $\epsilon = 10^{-2}$ are fixed.
With the starting point $x_0 = -2$ we reach Figure \ref{fig:Experiment2}.
In the magenta and the cyan area the ADAM method is not reaching the solution, although inequality \refb{eq:Kingma_Requirement} or \refb{eq:Our_Inequality} is satisfied. The ADAM is oscillating around the solution but do not reach them. The big difference is that the non-convergence in the cyan area is attributable to the fact that our proof only shows local convergence. By starting in $x_0 = -0.750000001$ the cyan area is almost complete blue (see figure \ref{fig:Experiment2_startvalue_near}). In contrary the magenta area does not change that much.
\subsection*{Experiment 3}
\begin{figure}[!t]
\center
\includegraphics[width=\bildgroesse , trim = 100 250 100 250, clip]{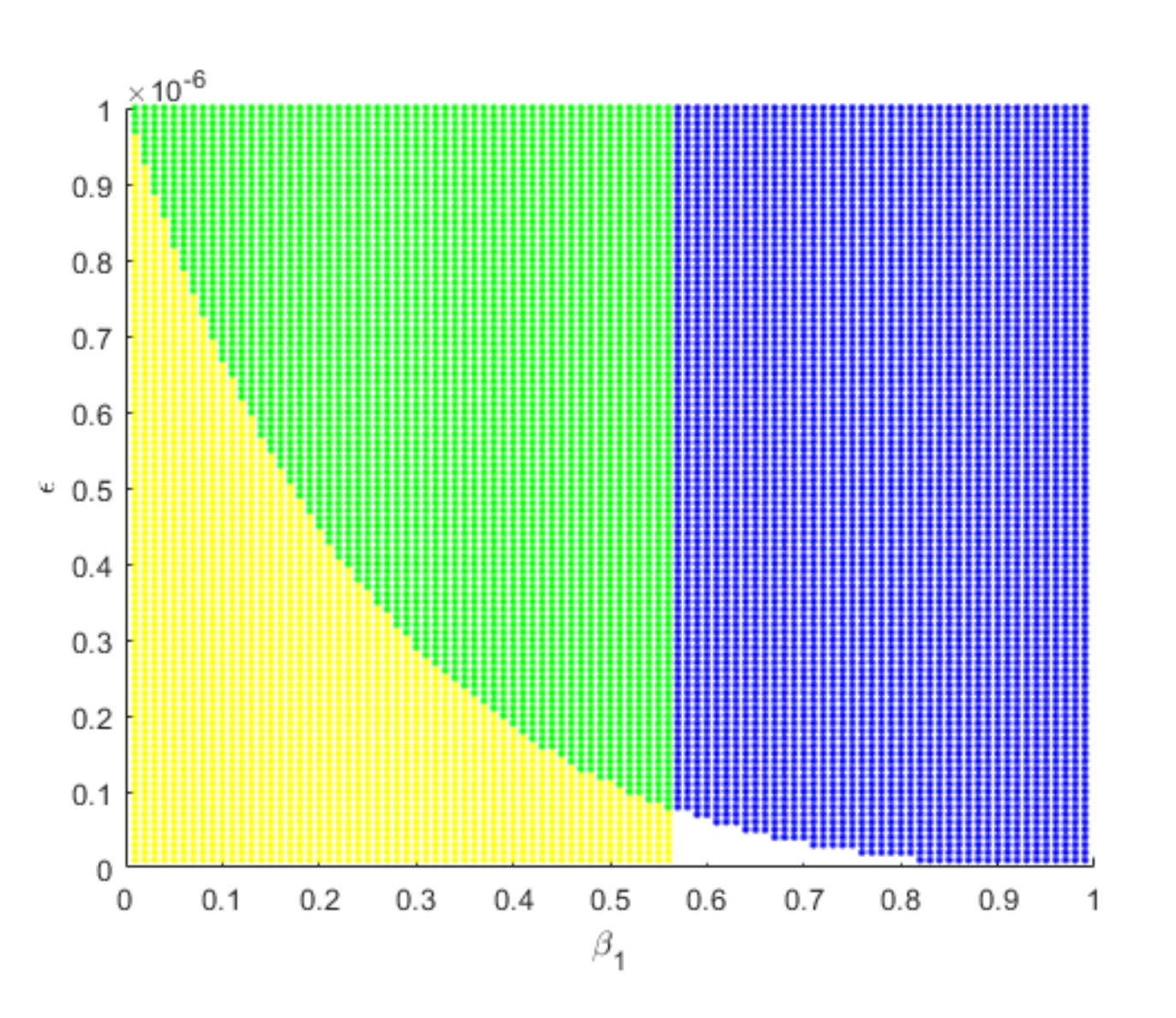}
\caption{Experiment 3 :Iterating over $\epsilon$ and $\beta_1$.}
\label{fig:Experiment_3}
\end{figure}

In the last experiment we use the same hyperparameters as in experiment $1$. Therefore we reach a similar looking Figure \ref{fig:Experiment_3} by iterating over the parameters.
The reason for the enlargement of the blue and green area is the different function $f \left( x \right)$, thus different eigenvalues in inequality \refb{eq:Our_Inequality}.
By observing the convergence behaviour from each of the four differently colored areas in figure \ref{fig:Experiment_3}, we can not spot big differences. 

\subsection{Experiments with other adaptive gradient decent optimizer}
\label{subsec:Experiments other optimizers}
The same experiments can be done with the optimizers of Table \ref{tab:optimizer extension hyperparameter bounding}. Since we only have dependencies on $\epsilon$ and $\alpha$, we only iterate over these hyperparameters with the same color coding (see Table \ref{tab:Colortable}) without any inequality beside our hyperparameter bounding. To avoid overloading the paper, the experiments can be found in the appendix.

However, AdaDelta plays a special role here, because it is quite astonishing that if you use the parameters suggested by Zeiler ($\learningrate = 1$) the convergence behaviour only depends on the eigenvalues of the Hessian of $f$. To be more clear, there are problems which converge or not converge unimpressed by the hyperparameters. To prove this theoretical result, we use the function $f(x)=\frac{1}{2} cw^2$ with $ \max \left(\mu\right) = c$. For $c=1.9$ -- our inequality satisfied -- we reach a convergent behaviour for each hyperparameter setting we tested. 
In contrast, if we choose $c=2.1$ -- our inequality is not satisfied -- we reach a non convergent behaviour for each hyperparameter setting. 
At this point we do not use pictures, because they would only show green or red dots.
%

\begin{figure}
\centering
\includegraphics[width=\bildgroesse , trim = 100 250 100 250, clip]{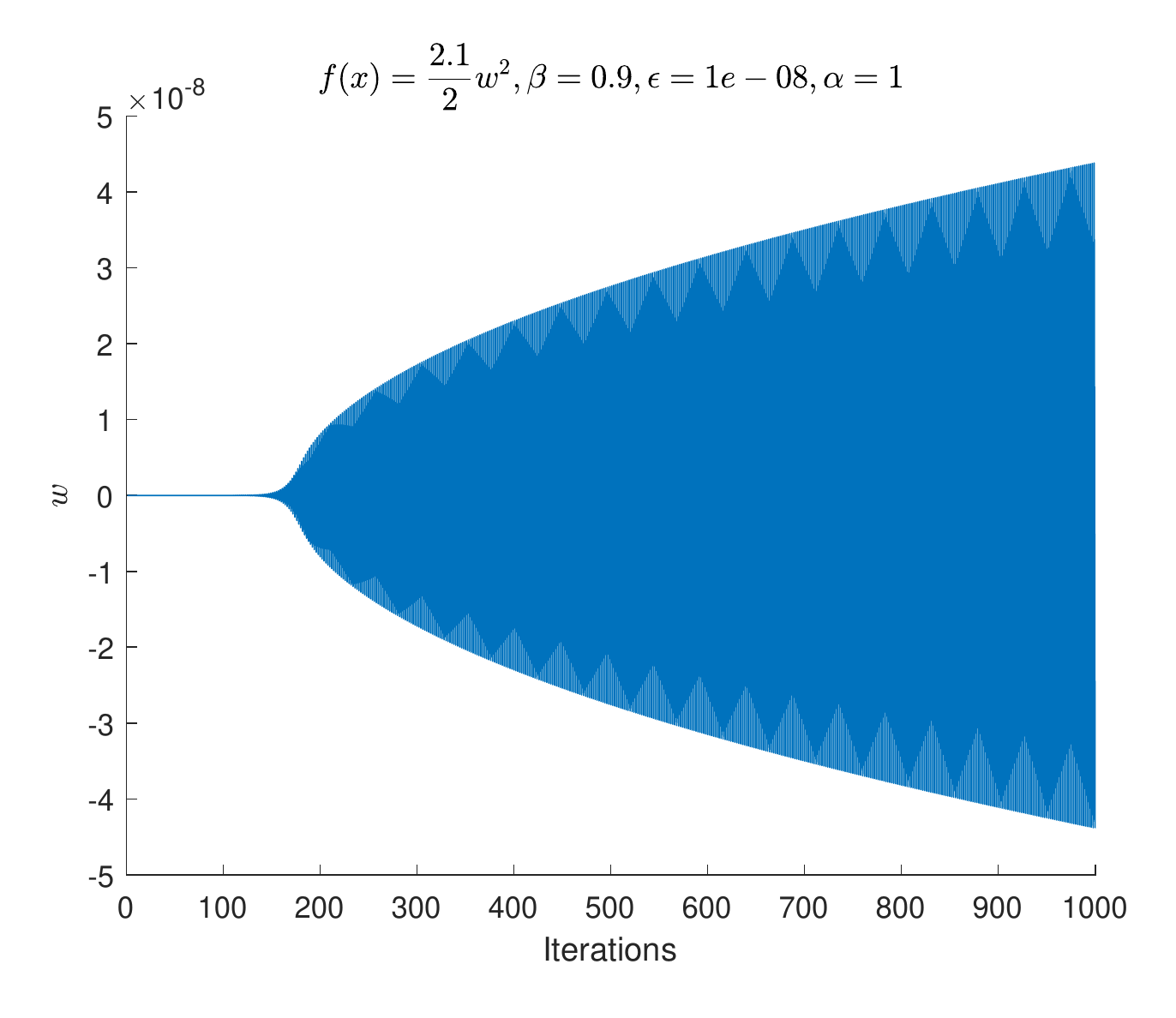}
\caption{Convergence behaviour with fixed hyperparameters}
\label{fig:AdaDelta-21-fixed-hyperparameters}
\end{figure}

\begin{figure}
\centering
\includegraphics[width=\bildgroesse , trim = 100 250 100 250, clip]{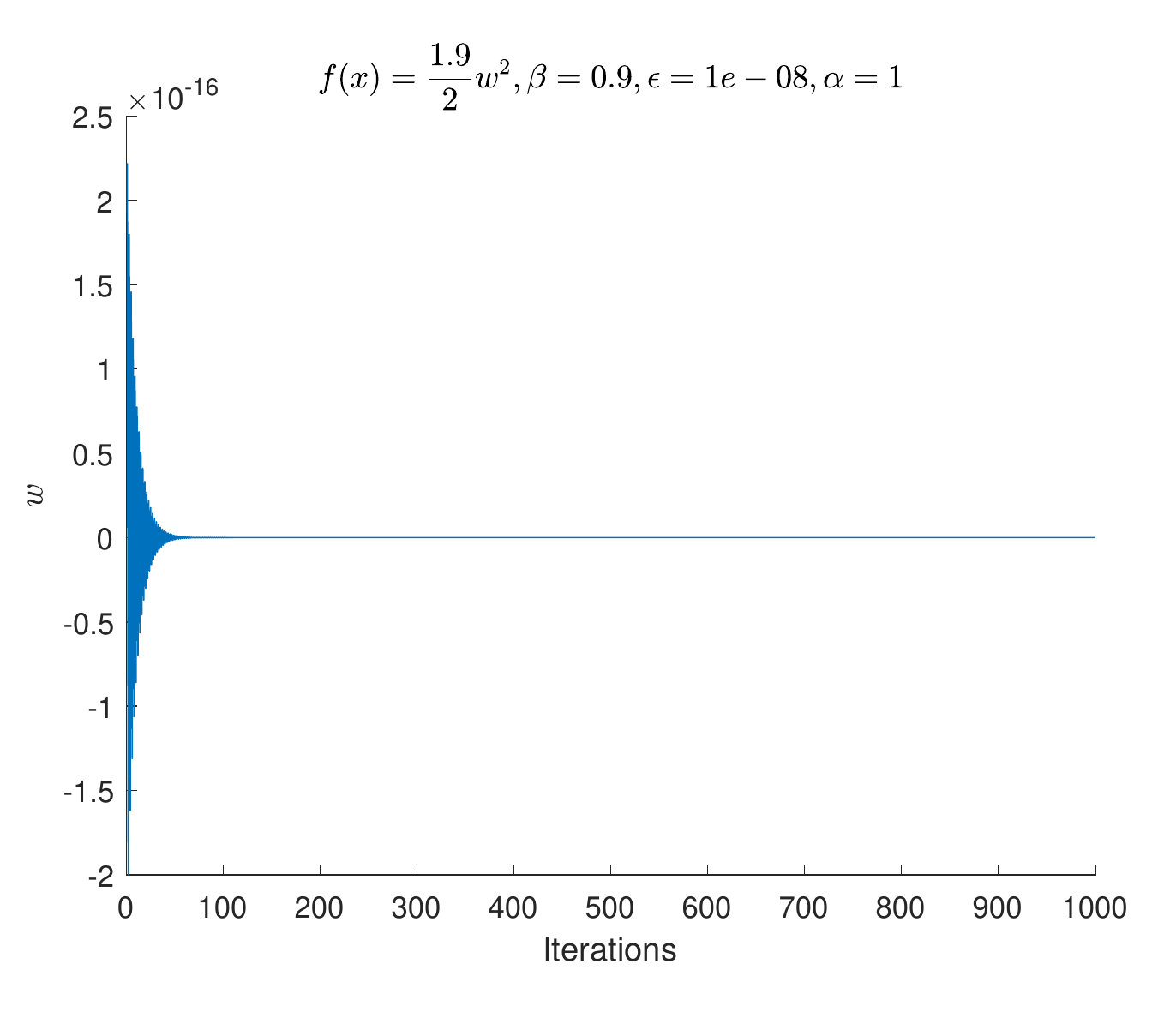}
\caption{Convergence behaviour with fixed hyperparameters}
\label{fig:AdaDelta-19-fixed-hyperparameters}
\end{figure}

If we look closer to some fixed hyperparameters we can see, that we are reaching the minimum in both cases. But with $c=2.1$ we are leaving the minimum at approximately iteration $200$ (compare Figure \ref{fig:AdaDelta-21-fixed-hyperparameters} and Figure \ref{fig:AdaDelta-19-fixed-hyperparameters}).

\begin{figure}
\centering
\includegraphics[width=\bildgroesse , trim = 100 250 100 250, clip]{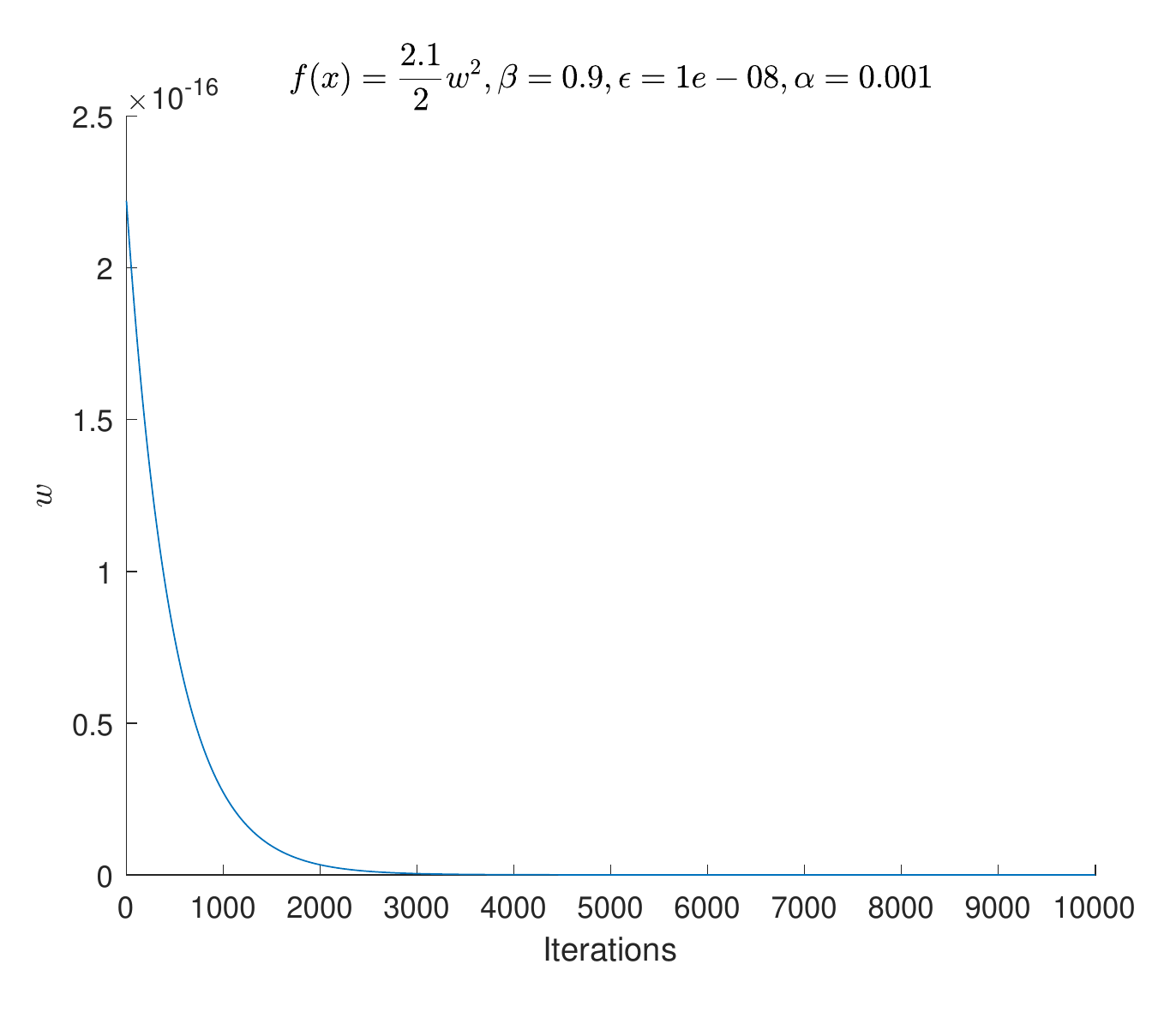}
\caption{Convergence behaviour with $\learningrate = 0.001$}
\label{fig:AdaDelta-21-with-learningrate}
\end{figure}

Different to the original paper from Zeiler \cite{Zeiler.2012}, commonly used frameworks for neural networks add a learning rate to the optimizer. Examples are the implementations in Tensorflow \cite{Abadi.2016} or in PyTorch \cite{Paszke.2019}. Tensorflow even goes one step further and mentions the original paper with $\learningrate = 1.0$ but sets the default value to $\learningrate = 0.001$. As previously shown, choosing a small learning rate can have a positive effect on convergence. See for example Figure \ref{fig:AdaDelta-21-with-learningrate} with $\learningrate=0.001$. Here we converge more slowly but we do not leave the minimum afterwards.

\begin{figure}
\centering
\includegraphics[width=\bildgroesse , trim = 100 250 100 250, clip]{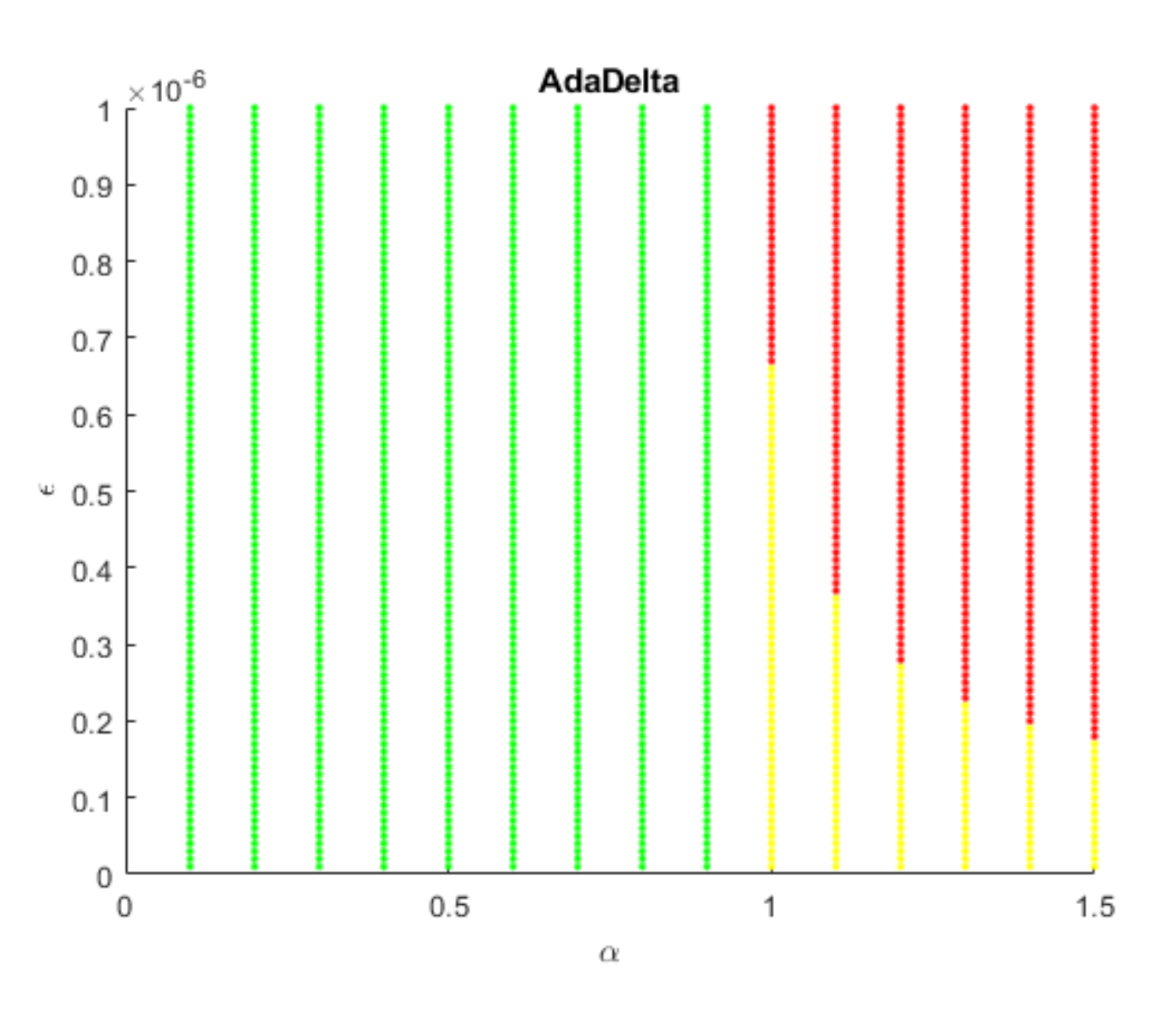}
\caption{Convergence behaviour by iterating over $\learningrate$ and $\epsilon$}
\label{fig:AdaDelta-21-iterating-over-alpha}
\end{figure}

By iterating over the learning rate $\learningrate$ and $\epsilon$ with $c=2.1$, we can prove again our convergence inequality with the colors coded by Table \ref{tab:Colortable} (see Figure \ref{fig:AdaDelta-21-iterating-over-alpha}). Even if we increase the number of iterations, the red area expands at the cost of the yellow area. 

\section{Conclusion and Discussion}
\label{sec:Conclusion and Discussion}
In this paper we have presented a local convergence proof of Algorithm \ref{alg:ADAM}, the original ADAM \cite{Kingma.2015}, SGD \cite{Robbins.1951}, RMSProp \cite{Hinton.2012} and AdaDelta \cite{Zeiler.2012}. To the best of our knowledge it is the first at all for the ADAM optimizer. We also give an a posteriori boundary for the hyperparameters and show, that the choice of $\beta_2$ does not matter for the convergence near a minimum. 

However the proof is based on the vanishing gradient condition $\nabla f(\weightstar)=0$
and  cannot be used for an incremental algorithm for 
$f(\weight) = \frac 1 N \sum_{i=1}^N f_i(\weight)$ where different component gradients
$g\left(w_t\right) = \nabla f_{i_t}(\weight_t)$ are used in the iterations for the moments. Clearly
$\nabla f(\weightstar)=0$ does not imply $\nabla f_i(\weightstar)=0$ for all components.
We are investigating how the incremental dynamical system can be related to the batch
system.

The analysis applies to any local minimum with positive definite Hessian
and therefore does not require overall convexity. 
In order to show global convergence of ADAM-like
algorithms other methods have to be applied.

\appendices
\section{Different $\epsilon$ positions}

The following two figures describes the accuracy and loss for the fashion-mnist dataset. The only different is the position of the $\epsilon$. 

\section{Convergence Proof}
\begin{proof}\textbf{(Theorem \ref{th:Eigenvalues})}\\
\label{pr:Eigenvalues Calculation}
We see that $J_{\bar{T}} \left( 0,0,\weightstar \right)$ has the $n$-fold eigenvalue 
$\beta_2$. So we can drop second block row and column of $J_{\bar{T}}$ and investigate the
eigenvalues of
\begin{align*}
\begin{bmatrix}
\beta_1 I & \left( 1-\beta_1\right) \nabla_w g\left( \weightstar \right)\\
-s \beta_1 I & I - s \left( 1-\beta_1\right) \nabla_w g \left( \weightstar \right)
\end{bmatrix}
=: 
\begin{bmatrix}
  A & B \\
  C & D
\end{bmatrix}
\end{align*}
where we use the abbreviation $s:=\frac{\learningrate}{\epsilon}$.
$B$ and $D$ are symmetric since $\nabla_w g \left( \weightstar \right)$ is the Hessian of $f$. 
By the spectral theorem we can diagonalize $B$ as 
$B = Q \Lambda Q^\transpose$ with an orthogonal matrix $Q$ and a diagonal matrix of
eigenvalues $\Lambda$. Analogously holds $D = I - Q \Lambda Q^\transpose = Q \left( I - \Lambda \right) Q^\transpose$.
We make a similarity transformation
with $\tilde{Q} := \begin{bmatrix}
Q & 0\\
0 & Q
\end{bmatrix} \in \Mat_{2n}$. 
This leaves the eigenvalues unchanged and gives 
\begin{align*}
\tilde{Q}^\transpose  
\begin{bmatrix}
  A & B \\
  C & D
\end{bmatrix}
\tilde{Q}
=\begin{bmatrix}
 \beta_1 I & \left( 1-\beta_1\right) \eigenvalueshessian_i I \\ -s \beta_1 I & I - s \left( 1 - \beta_1 \right) \eigenvalueshessian_i I
\end{bmatrix}
\end{align*}
with $\eigenvalueshessian_i$ the $i$-th eigenvalue of the Hessian. Eigenvalues does not change in similarity transformations, 
so we can also calculate the eigenvalues of our new block matrix with four diagonal sub matrices.
\begin{align*}
& \det \begin{bmatrix}
\left( \beta_1 - \lambda \right) I & \left( 1 - \beta_1 \right) \eigenvalueshessian_i I\\
-s \beta_1 I & \left(1-s\left(1-\beta_1\right)\eigenvalueshessian_i - \lambda \right) I
\end{bmatrix}\\
=& \det ( \left( \beta_1 - \lambda \right) \left( 1-s\left(1-\beta_1\right)\eigenvalueshessian_i - \lambda \right) I\\ 
&+ \left( 1-\beta_1 \right) s \beta_1 \eigenvalueshessian_i I )\\
\overset{!}{=}&0
\end{align*}
Therefore the matrix is a diagonal matrix, we can conclude:
\begin{align*}
\det \left( J_{\bar{T}} \left( \xstar \right) - \lambda I \right) =& \prod_{i=1}^n \left( \beta_1 - \lambda \right) \left( 1 - s\left(1-\beta_1\right)\eigenvalueshessian_i - \lambda \right)\\
&+ \left( 1 - \beta_1 \right) s \beta_1 \eigenvalueshessian_i
\end{align*}
Each factor can be written as
\begin{align*}
\lambda^2 - \left( 1- s \left( 1-\beta_1\right) \eigenvalueshessian_i +\beta_1\right) \lambda + \beta_1 &\overset{!}{=} 0
\end{align*} 
and following the statement
\begin{align*}
\lambda_{23,i} =& \frac 1 2 \left( 1-s\left(1-\beta_1 \right) \eigenvalueshessian_i + \beta_1 \right. \\ 
&\left. \pm \sqrt{\left( 1-s\left(1-\beta_1\right) \eigenvalueshessian_i + \beta_1 \right)^2 - 4\beta_1} \right)
\end{align*}
is true.
\end{proof}
\begin{proof}\textbf{(Theorem \ref{th:Spectral-radius-Jacobian})}\\
\label{pr:Spectral radius eigenvalues}
We already have calculate the eigenvalues of the Jacobian in Theorem \ref{th:Eigenvalues}. With these we can easily see, that $|\lambda_1| = |\beta_2| < 1 $ is satisfied per the requirements of algorithm \ref{alg:ADAM}. Therefore we define $\varphi_i := \frac{\learningrate \eigenvalueshessian_i}{\epsilon} \left( 1-\beta_1 \right)$ and observe the absolute value of the two eigenvalues left.
\begin{align*}
|\lambda_{23,i}| = \frac{1}{2} \left| \left( 1 + \beta_1 - \varphi_i \right) \pm \underbrace{\sqrt{ \left(1 + \beta_1 - \varphi_i \right)^2 - 4 \beta_1}}_{\text{\circone}} \right|
\end{align*}
First we look at upper bound of the eigenvalues. For this we take term \circone $ $ combined with the regrets for $\varphi_i$:
\begin{align*}
\sqrt{\left(1 + \beta_1 - \varphi_i \right)^2 - 4 \beta_1} <& \sqrt{\left(1 + \beta_1 \right)^2 - 4 \beta_1}\\ 
=& \pm \left(1 - \beta_1 \right)
\end{align*}
So if we put this in $\lambda_{23,i}$ we have the inequality $\left|\lambda_{23,i}\right| < \frac{1}{2} \left| 1+\beta_1-\varphi_i \pm \left(1-\beta_1\right) \right|$. Easy to see are the two cases:
\begin{align*}
\left|\lambda_{23,i}\right| &< 1 &&\text{with +}\\
\left| \lambda_{23,i} \right| &< \beta_1 < 1 &&\text{with -}
\end{align*}
In both cases we see that the eigenvalues are smaller than $1$ in absolute value. 
To show the lower bound $\lambda_{23,i} > -1$, we look again at term \circone.
\begin{align*}
\underbrace{\sqrt{\left(1+\beta_1 - \varphi_i \right)^2 - 4 \beta_1}}_{\in \C\backslash \R} = i \sqrt{4 \beta_1 - \left(1+\beta_1 - \varphi_i \right)^2}
\end{align*}
Then we can write:
\begin{align*}
|\lambda_{23,i}| =& \frac{1}{2} \sqrt{\left(1+\beta_1 - \varphi_i \right)^2 + 4 \beta_1 - \left(1+\beta_1-\varphi_i \right)^2 }\\
=& \sqrt{\beta_1} < 1
\end{align*}
The last inequality is given by the requirements of Theorem \ref{th:Spectral-radius-Jacobian} and so we proved the whole Theorem.
\end{proof}





\begin{theorem} 
\label{th:ConvergencePerturbation}
\textbf{Convergence to fixed point with perturbation}
Let $M\subset \R^n$ be a complete set, $\bar T: M \to M$ Lipschitz continuous with $L<1$, 
 $\xstar\in M$ the unique fixed point of $\bar{T}$.
Assume $B_r(\xstar) \subset M$ for
some $r>0$. Recall that the non-autonomous system \refb{eq:System_Origin} is defined by
\begin{equation*}
	\tilde x_{t+1} = T(\tilde x_t) := \bar T(\tilde x_t) + \Theta(t,\tilde x_t)
\end{equation*}
for $\Theta: \N_0\times M \to\R^n$ 
with the bound $\norm{\Theta(t,\tilde x)} \leq C \beta^t \norm {\tilde x-\xstar}$
for all $\tilde x_t\in M$, $t\in \N_0$ for some $C\geq 0$ and $0< \beta < 1$. 
Then there exists $\varepsilon>0$ such that for all $\tilde x_0\in M$ with 
$\norm{\tilde x_0-\xstar}<\varepsilon$ the iteration defined by 
\refb{eq:System_Origin} is well-defined, i.e. stays in $M$, and converges to
$\xstar$.
\end{theorem}
\begin{proof}
Let $x = x(\cdot, \tilde x_0)$ be the solution of the undisturbed iteration 
$x_{t+1} = \bar T(x_t)$ with
initial condition $\tilde x_0$, $\tilde x = \tilde x(\cdot, \tilde x_0)$ the
corresponding solution of \refb{eq:System_Origin}. 
We define $e_t:=\norm{\tilde x_t - \xstar}$, and estimate using the assumptions
\begin{eqnarray*}
	e_{t+1} 
	&=& 
	\norm {\bar T(\tilde x_t) + \Theta(t,\tilde x_t) - \xstar}
	\\
	&=& 
	\norm {\bar T(\tilde x_t) - \bar T (\xstar) + \Theta(t,\tilde x_t) }
	\\
	&\leq& 
	\norm {\bar T(\tilde x_t) - \bar T(\xstar)} + \norm{\Theta(t,\tilde x_t)}
	\\
	&\leq& 
	L \norm{\tilde x_t- \xstar} +  C \beta^t \norm {\tilde x_t-\xstar}
	\\
	&=& 
	(L + C \beta^t) e_t 
\end{eqnarray*}
Choosing $t$ large enough, we get $0 < L +  C \beta^t\leq \tilde L < 1$
for all $t \geq K$ because $\beta, L<1$. 
Then
\begin{eqnarray*}
	e_t \leq & \left(\prod_{k=1}^K (L + C \beta^k) \right) \tilde L^{t-K} e_0 =: & \tilde C \tilde L^{t-K} e_0
\end{eqnarray*}
with $\tilde C$ independent of $\tilde x_0$. 
So $e_t$ converges to 0 exponentially.

The arguments so far have only been valid if $\tilde x_t\in M$, i.e. the 
iteration is well defined. But choosing $\tilde x_0$ such that 
$e_0=\norm{\tilde x_0 - \xstar} < \frac r {\tilde C} $
small enough that we can achieve $e_t \leq \tilde C e_0 < r$. 
\end{proof}

\begin{theorem} \textbf{Determinants of Block Matrices \cite{Silvester.1999}}\\ 
\label{th:Silvester-Theorem}
Let $M = \begin{bmatrix}
A B\\
C D
\end{bmatrix} \in \Mat_{2n}$ be a block matrix with $A, B, C, D \in \R^{n \times n}$. If $C$ and $D$ commute, then $\det\left( M \right) = \det \left( AD-BC \right)$ holds.
\end{theorem}



\begin{lemma} 
\label{lemmaGradientToF}
Let $f\in C^2(\R^n, \R)$, $\xstar\in\R^n$ with $\nabla f(\xstar)=0$ and $\nabla^2 f(\xstar)$ invertible.
Then there exist $\epsilon>0$ and $C>0$ with $\norm{\nabla f(x)} \geq C \norm{x-\xstar}$ for all
$x\in\ball \epsilon \xstar$.
\end{lemma}
\begin{proof}
As $f$ is $C^2$ we have 
$\nabla f(x) - \nabla f(\xstar) - \nabla^2 f(\xstar) = o(\norm {x-\xstar})$. So for each
$\delta>0$ there exists $\epsilon>0$ with 
$\norm{\nabla f(x) - \nabla f(\xstar) - \nabla^2 f(\xstar)} \leq \delta \norm {x-\xstar}$
for all $x\in\ball \epsilon \xstar$. Assume w.l.o.g. $\delta < \frac 1 {\norm {\nabla^2 f(\xstar)^{-1}}}$.
Then we have 
\begin{align*}
	\norm{\nabla f(x)} & 
	\geq
	\norm{\nabla^2 f(\xstar) (x-\xstar)} \\
	& \quad - \norm{ \nabla f(x)-\nabla f(\xstar) -\nabla^2 f(\xstar) (x-\xstar)}
	\\
	&\geq 
	\frac 1 {\norm {\nabla^2 f(\xstar)^{-1}}} \norm{x-\xstar} - \delta \norm{x-\xstar}
	\\
	& =: C \norm{x-\xstar}
\end{align*}
with $C=\frac 1 {\norm {\nabla^2 f(\xstar)^{-1}}} - \delta > 0$ by choice of $\delta$.
\end{proof}

\begin{theorem}\textbf{Exponential stability implies existence of a Lyapunov function, arbitrary fixed point}\\
\label{th:Bof_27xstar}
Let $\xstar$ be an fixed point for the nonlinear system the autonomous system
$
x(t+1) = f\left(x\left(t\right) \right)
$
where $f:D \rightarrow \R^n$ is continuously differentiable and $D=\{ x\in \R^n | \norm{x-\xstar}<r \}$. Let $k, \lambda$ and $r_0$ be positive constants with $r_0 < r/k$. Let $D_0 = \{x \in \R^n | \norm{x-\xstar} < r_0 \}$. Assume that the solution of the system satisfy
\begin{align}
\label{eq:exponential_stability_xstar}
\norm{x(t, x_0)} \leq k \norm{x_0-\xstar} e^{-\lambda t}, \forall x_0 \in D_0, \forall t \geq 0
\end{align}
Then there there is a function $V: D_0 \rightarrow \R$ with
\begin{align*}
c_1 \norm{x-\xstar}^2 &\leq V\left(x\right)\leq c_2 \norm{x-\xstar}^2\\
V\left(f \left(x \right) \right) - V\left(x \right) &\leq -c_3 \norm{x-\xstar}^2\\
|V\left(x\right) - V\left(y \right)| &\leq c_4 \norm{x-y} \left(\norm{x-\xstar}+\norm{y-\xstar}\right)
\end{align*}
\end{theorem}


Note: In the proof of this theorem the norm denotes the 2-norm. The equivalence of norms
then easily transfers the result to other norms. 
\begin{proof}
Let $\phi \left(t,x_0\right)$ be the solution of $x_{t+1} = f\left(x_k\right)$ at time $t$ starting from $x_0$ at time $k=0$ and $\xstar$ be an equilibrium point of the system. Let
\begin{align*}
V\left(x_0\right) = \sum\limits_{t=0}^{N-1} (\phi \left(t,x_0\right) - \xstar )^\transpose \left(\phi \left(t,x_0\right) -\xstar\right)
\end{align*}
for some integer variable $N$ to be set. Then 
\begin{align*}
V\left(x_0\right) =& \left(x_0 - \xstar \right)^\transpose \left(x_0-\xstar\right)\\
&+\sum\limits_{t=1}^{N-1} (\phi \left(t,x_0\right) - \xstar )^\transpose \left(\phi \left(t,x_0\right) -\xstar\right) \\
\geq& \left(x_0-\xstar\right)^\transpose \left( x_0-\xstar \right) = \norm{x_0-\xstar}
\end{align*}
and on the other hand, using \refb{eq:exponential_stability_xstar} we have 
\begin{align*}
V\left(x_0\right) &= \sum\limits_{t=0}^{N-1} \left( x_t - \xstar \right)^\transpose \left( x_t -\xstar\right)\\ 
&\leq \sum\limits_{t=0}^{N-1} k^2 \norm{x_0-\xstar}^2 e^{-2\lambda t}\\ 
&\leq k^2 \left(\frac{1-e^{-2\lambda N}}{1-e^{-2 \lambda}} \right) \norm{x_0 - \xstar}^2
\end{align*}
We have shown that there exists $c_1$ and $c_2$ such that
\begin{align*}
c_1 \norm{x_0-\xstar}^2 \leq V \left(x_0\right) \leq c_2 \norm{x_0-\xstar}^2
\end{align*}
is satisfied. Now, since $\phi \left( t,f\left( x_0\right)\right)= \phi \left(t,\phi \left(1,x_0\right) \right) = \phi \left(t+1,x_0\right)$,
\begin{align*}
&V\left(f\left(x\right)\right) - V\left(x\right)\\ 
=& \sum\limits_{t=0}^{N-1} \left( \phi \left(t+1,x_0\right)-\xstar\right)^\transpose \left(\phi \left(t+1,x_0\right)-\xstar \right) \\
&- \sum\limits_{t=0}^{N-1} \left(\phi \left(t,x_0\right)-\xstar\right)^\transpose \left(\phi \left(t,x_0\right)-\xstar\right)\\
=& \sum\limits_{j=1}^{N} \left(\phi \left(j,x_0\right)-\xstar \right)^\transpose \left(\phi \left(j,x_0\right)-\xstar \right)\\ 
&- \sum\limits_{t=0}^{N-1} \left(\phi \left(t,x_0\right) -\xstar \right)^\transpose \left(\phi\left( t,x_0 \right)-\xstar \right)\\ 
=& \left(\phi \left(N,x_0\right)-\xstar \right)^\transpose \left(\phi \left(N,x_0\right)-\xstar \right)\\ 
&- \left(x_0 -\xstar\right)^\transpose \left(x_0-\xstar\right)\\
\leq& k^2 e^{-2\lambda n} \norm{x_0-\xstar}^2 - \norm{x_0-\xstar}^2\\
=& - \left( 1-k^2 e^{-2 \lambda N} \right) \norm{x_0-\xstar}^2
\end{align*}
Now we choose $N$ big enough so that $1-k^2 e^{-2 \lambda N}$ is greater than $0$ and also the second property has been proven. For the third property, since $f$ is continuously differentiable it is also Lipschitz over the bounded domain $D$, with a Lipschitz constant $L$, for which it holds $\norm{f\left(x\right) - f\left(y \right)} \leq L \norm{x-y}$. Then 
\begin{align*}
&\norm{\phi\left(t+1,x_0\right)-\phi\left(t+1,y_0\right)}\\
=& \norm{f\left(\phi\left(t,x_0\right)\right) - f\left(\phi\left(t,y_0\right)\right)}\\ 
\leq& L \norm{\phi\left(t,x_0\right) - \phi\left(t,y_0\right)}
\end{align*}
and by induction
\begin{align*}
\norm{\phi\left(t,x_0\right) - \phi \left(t,y_0\right)} \leq L^t \norm{x_0-y_0}
\end{align*}
Consider now the difference $|V\left(x_0\right) - V\left(y_0\right)|$
\begin{align*}
=& \big| \sum\limits_{t=1}^{N-1} \Big( \left(\phi \left(t,x_0\right)-\xstar\right)^\transpose \left(\phi \left(t,x_0\right)-\xstar\right)\\ 
&- \left(\phi \left(t,y_0\right)-\xstar \right)^\transpose \left(\phi \left(t,y_0 \right) -\xstar \right) \big) \Big|\\
=& \Big| \sum\limits_{t=0}^{N-1} \big(\left(\phi \left(t,x_0 \right)-\xstar\right)^\transpose \\ 
&\cdot \left(\left(\phi \left(t,x_0\right)-\xstar\right)- \left(\phi \left(t,y_0\right)-\xstar\right)\right) \\
&+ \left(\phi \left(t,y_0\right)-\xstar\right)^\transpose \\
&\cdot \left(\left( \phi \left(t,x_0\right)-\xstar\right)-\left(\phi\left(t,y_0 \right)-\xstar\right)\right)\big)\Big|\\
\leq& \sum\limits_{t=0}^{N-1} \big( \norm{\phi \left(t,x_0\right)-\xstar} \norm{\phi\left(t,x_0\right)-\phi\left(t,y_0\right)}\\
&+ \norm{\phi  \left(t,y_0\right)-\xstar} \norm{\phi\left(t,x_0\right)-\phi\left(t,y_0\right)} \big)\\
\leq& \sum\limits_{t=0}^{N-1} \left( \norm{\phi \left(t,x_0\right)-\xstar} + \norm{\phi \left(t,y_0\right)-\xstar}\right) L^t \norm{x_0-y_0}\\
\leq& \left[\sum\limits_{t=0}^{N-1} k e^{-\lambda t} L^t \right] \left(\norm{x_0-\xstar} + \norm{y_0-\xstar} \right) \norm{x_0-y_0}\\
\leq& c_4 \left(\norm{x_0-\xstar}+\norm{y_0-\xstar}\right)\norm{x_0-y_0}
\end{align*}
and so we have proven the last inequality.
\end{proof}

\begin{theorem}\textbf{Exponential convergence under disturbances}
\label{th:ExponentialConvergenceUnderDisturbances}
Suppose $f:\R^n\to\R^n$ Lipschitz continuous with $L_f$ and the discrete time system
$
x_{t+1}=f(x_t), \quad x(0)=x_0  
$
 Assume $f(\xstar)=\xstar$ and that 
 $\xstar$ is a global exponentially stable fixed point. 
Assume $h:\R^n\to\R^n$ Lipschitz continuous with $L_h$, with $h(\xstar)=0$ and 
$\norm {h(x)} \leq L_h \norm {x-\xstar}$ for all $x\in\R^n$ 
and the discrete time system
\begin{equation}
\label{eqExpStableDTdisturbance}
\tilde x_{t+1}=f(\tilde x_t)+h(\tilde x_t),  \quad \tilde x(0)=\tilde x_0
\end{equation}
Then, if $L_h$ is small enough, $\tilde{x}_\star=\xstar$
is a global exponentially stable fixed point for (\ref{eqExpStableDTdisturbance}) as well.
\end{theorem}
\begin{proof} 
The converse Lyapunov theorem ensures the existence of a function $V$ with properties
\begin{align*}
c_1 \norm{x-\xstar}^2 &\leq V\left(x\right)\leq c_2 \norm{x-\xstar}^2\\
\Delta V(x) &= V\left(x_{t+1}\right) - V \left(t\right) \leq -c_3 \norm{x-\xstar}^2\\
|V\left(x\right) - V\left(y \right)| &\leq c_4 \norm{x-y} \left(\norm{x-\xstar}+\norm{y-\xstar}\right)
\end{align*}
We show that, for $L_h$ small enough, $\tilde V(\tilde x):=V(\tilde x)$ 
is also a Lyapunov function for 
(\ref{eqExpStableDTdisturbance}). We use the tilde symbol to denote time derivatives
along (\ref{eqExpStableDTdisturbance}):
\begin{align*}
\tilde{\Delta} \tilde{V} \left(\tilde{x} \right) =& \tilde{V} \left(f\left(\tilde{x}\right)+h\left(\tilde{x}\right)\right)-\tilde{V}\left(\tilde{x}\right)\\
=& V\left(f\left(\tilde{x}\right)+h\left(\tilde{x}\right)\right)-V\left(\tilde{x}\right)\\
=& V\left(f\left(\tilde{x}\right)\right) -V\left(\tilde{x}\right)+V\left(f\left(\tilde{x}\right)+h\left(\tilde{x}\right)\right)\\
&-V\left(f\left(\tilde{x}\right)\right)
\end{align*}
The properties of $V$ and $h$ give
\begin{align*}
\tilde{\Delta}\tilde{V}\left(\tilde{x}\right) \leq& -c_3 \norm{\tilde{x}-\xstar}^2 + c_4 \norm{h\left(\tilde{x}\right)}\\
&\cdot \left(\norm{f\left(\tilde{x}\right)+h\left(\tilde{x}\right)-\xstar}+\norm{f\left(\tilde{x}\right)-\xstar}\right)\\
=& -c_3 \norm{\tilde{x}-\xstar}^2 + c_4 \norm{h\left(\tilde{x}\right)}\\
&\cdot \left(\norm{f\left(\tilde{x}\right) - \xstar + h\left(\tilde{x}\right)} + \norm{f\left(\tilde{x}\right) - \xstar}\right)\\
\leq& -c_3 \norm{\tilde{x}-\xstar}^2 + c_4 L_h \norm{\tilde{x}-\xstar}\\
&\cdot (\norm{f\left(\tilde{x}\right)-f\left(\xstar\right)} + \norm{h\left(\tilde{x}\right)} \\
&+ \norm{f\left(\tilde{x}\right)-f\left(\xstar\right)})\\
\leq& -c_3\norm{\tilde{x}-\xstar}^2 + c_4 L_h \norm{\tilde{x}-\xstar}\\
&\cdot \left(L_f \norm{\tilde{x}-\xstar} + L_h \norm{\tilde{x}-\xstar} + L_f \norm{\tilde{x}-\xstar} \right)\\
=& - \left(c_3 - c_4 L_h \left(L_f + L_h + L_f \right) \right) \norm{\tilde{x}-\xstar}^2
\end{align*}

So for $L_h$ small enough we get exponential convergence.
\end{proof}



%
%


\section{Experiments for SGD and RMSProp}
\begin{figure}
\center
\includegraphics[width=\bildgroesse , trim = 100 250 100 250, clip]{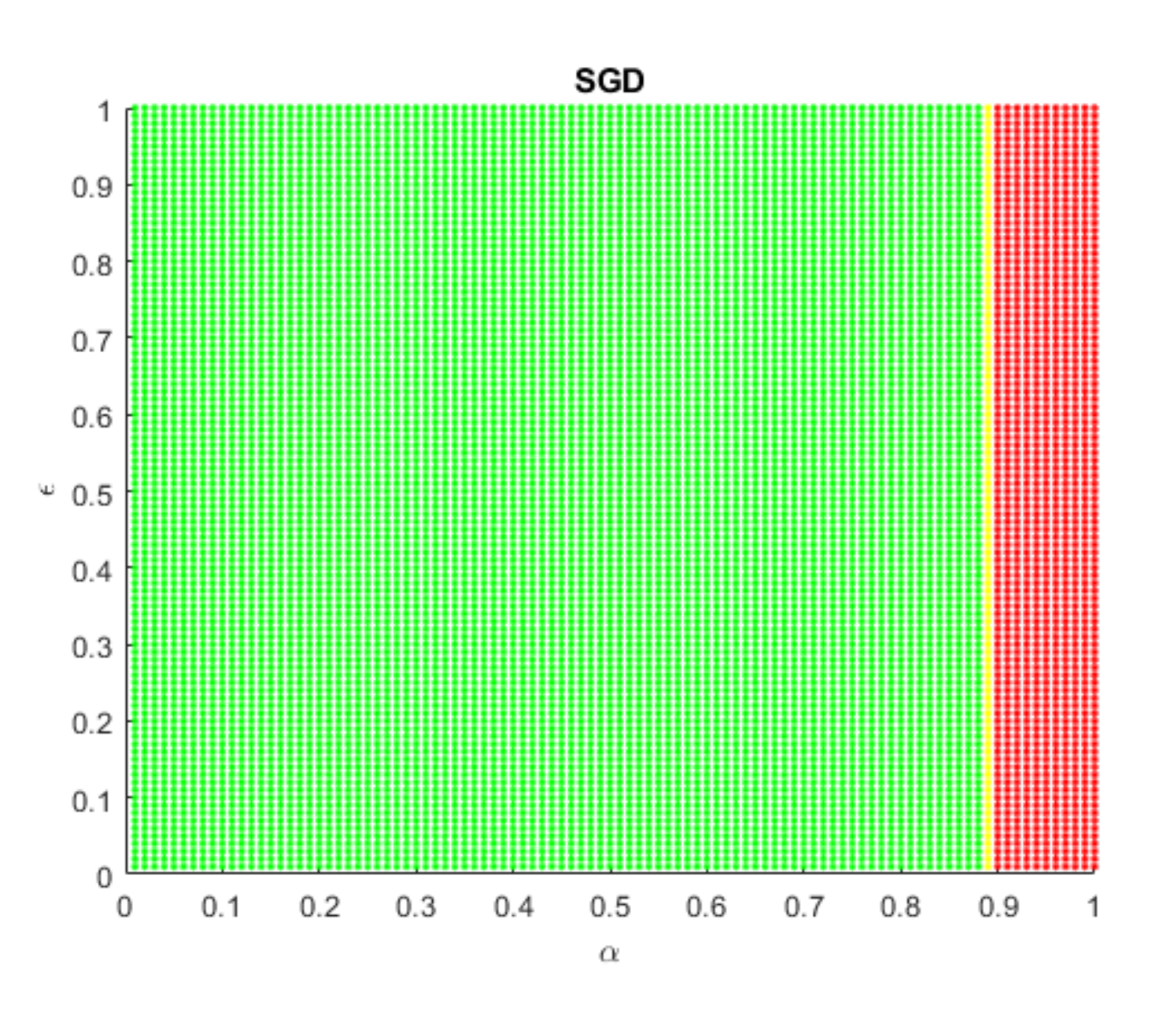}
\caption{Iterating over $\epsilon$ and $\learningrate$ with $x_0 = -0.65$}
\label{fig:Experiment 4}
\end{figure}

\begin{figure}
\center
\includegraphics[width=\bildgroesse , trim = 100 250 100 250, clip]{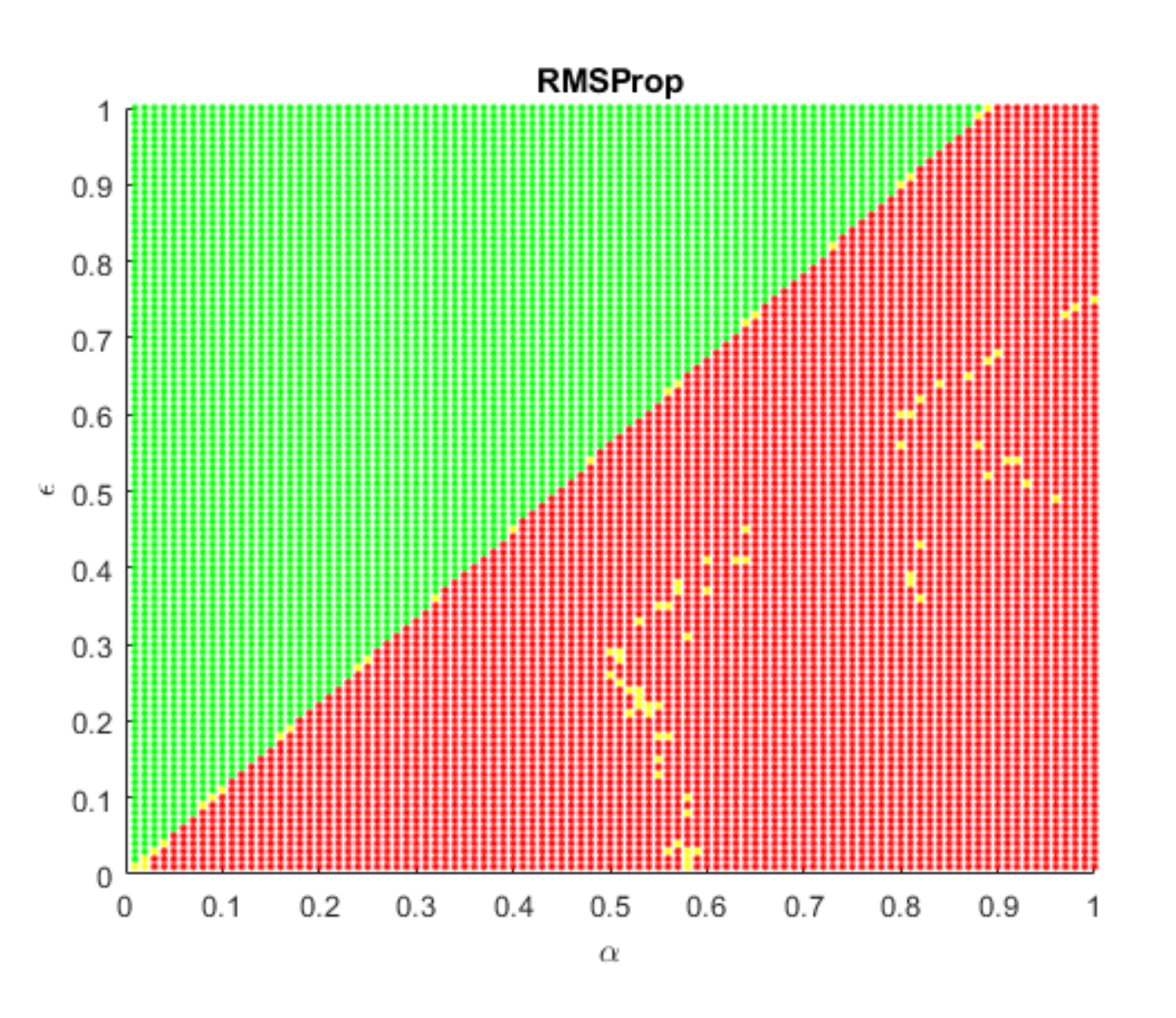}
\caption{Iterating over $\epsilon$ and $\learningrate$ with $x_0 = -2$}
\label{fig:Experiment 5}
\end{figure}

To prove the validity of the inequalities in Table \ref{tab:optimizer extension hyperparameter bounding}, we make the same experiments as in Subsection \ref{subsec:Solution Behaviour}. We set the fixed hyperparameter $\beta = 0.1$ and iterate over $\epsilon \in \{10^{-2}, \cdots,  1\}$ and $\learningrate \in \{0.01, \cdots, 1\}$. In Figure \ref{fig:Experiment 4} we can see our predicted vertical border at $\learningrate = 0.88$ and after a short yellow area -- where the SGD also converge but our hyperparameter bounding is not fulfilled -- we see the red area -- SGD is not converging -- at around $\learningrate = 0.9$.
In Figure \ref{fig:Experiment 5} we can see a similar behaviour. We can detect some yellow spots, where our inequality is not fulfilled but the RMSProp converges. However, there are no black areas where this is reversed. Denote that if we set $x_0$ far away from $x_\star$ we get some black area due to the fact that we only prove local convergence. In the experiments the SGD had a much smaller convergence area as RMSProp.

\section*{Acknowledgment}
This paper presents results of the project "LeaP – Learning Poses" supported by the Bavarian Ministry of Science and Art under Kap. 15 49 TG 78.

\ifCLASSOPTIONcaptionsoff
  \newpage
\fi

\end{document}